\documentclass{article} 
\usepackage{iclr2024_conference,times}

\usepackage{amsmath,amsfonts,bm}

\def\eqref#1{equation~\ref{#1}}

\def\1{\bm{1}}

\DeclareMathAlphabet{\mathsfit}{\encodingdefault}{\sfdefault}{m}{sl}
\SetMathAlphabet{\mathsfit}{bold}{\encodingdefault}{\sfdefault}{bx}{n}

\usepackage{hyperref}
\usepackage{url}
\usepackage{ulem}

\title{AnyText: Multilingual Visual Text Generation and Editing}

\author{Yuxiang Tuo, Wangmeng Xiang, Jun-Yan He, Yifeng Geng\thanks{corresponding author} , Xuansong Xie \\
Institute for Intelligent Computing, Alibaba Group\\
\texttt{\{yuxiang.tyx,wangmeng.xwm,leyuan.hjy,cangyu.gyf,xingtong.xxs\}}  \\
\texttt{@alibaba-inc.com}}

\newcommand{\xwmadd}[1]{\textcolor{blue}{#1}}

\iclrfinalcopy 
\usepackage{graphicx}
\usepackage{hyperref}
\usepackage{multirow}
\usepackage{verbatim}
\graphicspath{{./latex/fig/}}
\begin{document}

\maketitle

\begin{abstract}

Diffusion model based Text-to-Image has achieved impressive achievements recently. Although current technology for synthesizing images is highly advanced and capable of generating images with high fidelity, it is still possible to give the show away when focusing on the text area in the generated image, as synthesized text often contains blurred, unreadable, or incorrect characters, making visual text generation one of the most challenging issues in this field. To address this issue, we introduce \textbf{AnyText}, a diffusion-based multilingual visual text generation and editing model, that focuses on rendering accurate and coherent text in the image. AnyText comprises a diffusion pipeline with two primary elements: an auxiliary latent module and a text embedding module. The former uses inputs like text glyph, position, and masked image to generate latent features for text generation or editing. The latter employs an OCR model for encoding stroke data as embeddings, which blend with image caption embeddings from the tokenizer to generate texts that seamlessly integrate with the background. We employed text-control diffusion loss and text perceptual loss for training to further enhance writing accuracy. AnyText can write characters in multiple languages, to the best of our knowledge, this is the first work to address multilingual visual text generation. It is worth mentioning that AnyText can be plugged into existing diffusion models from the community for rendering or editing text accurately. After conducting extensive evaluation experiments, our method has outperformed all other approaches by a significant margin.
Additionally, we contribute the first large-scale multilingual text images dataset, \textbf{AnyWord-3M}, containing 3 million image-text pairs with OCR annotations in multiple languages. Based on AnyWord-3M dataset, we propose AnyText-benchmark for the evaluation of visual text generation accuracy and quality. Our project will be open-sourced soon
on \xwmadd{\url{https://github.com/tyxsspa/AnyText}} 
to improve and promote the development of text generation technology. 

\end{abstract}
\section{Introduction}
\begin{figure}[htbp]
 \centering
 \includegraphics[width=1.0\textwidth]{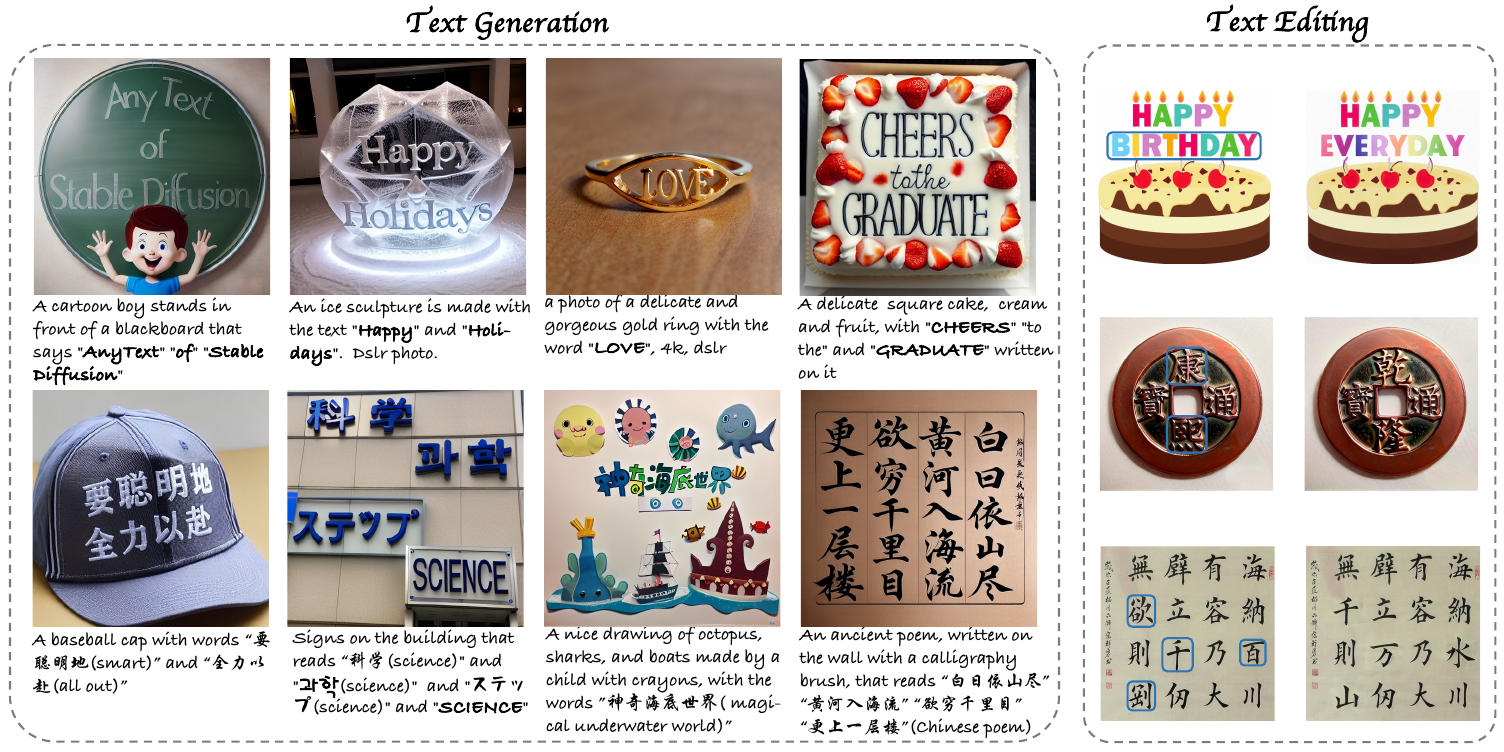}
 \caption{Selected samples of AnyText. For text generation, AnyText can render the specified text from the prompt onto the designated position, and generate visually appealing images. As for text editing, AnyText can modify the text content at the specified position within the input image while maintaining consistency with the surrounding text style. Translations are provided in parentheses for non-English words in prompt, blue boxes indicate positions for text editing. See more in ~\ref{app: more_examples}.}
 \label{fig:samples}
\end{figure}

Diffusion-based generative models~\cite{Saharia_Imagen_NIPS22, Ramesh_DALLE2_corr22, Rombach_LDM_CVPR22, Zhang_ControlNet_Corr23} demonstrated exceptional outcomes with unparalleled fidelity, adaptability, and versatility. Open-sourced image generation models (for example, Stable Diffusion~\cite{Rombach_LDM_CVPR22}, DeepFloyd-IF~\cite{DeepFloyd_23}), along with commercial services (Midjourney~\cite{midjourney}, DALL-E 2~\cite{Ramesh_DALLE2_corr22}, etc.) have made substantial impacts in various sectors including photography, digital arts, gaming, advertising, and film production. Despite substantial progress in image quality of generative diffusion models, most current open-sourced models and commercial services struggle to produce well-formed, legible, and readable visual text. Consequently, this reduces their overall utility and hampers potential applications.

The subpar performance of present open-source diffusion-based models can be attributed to a number of factors. Firstly, there is a deficiency in large-scale image-text paired data which includes comprehensive annotations for textual content. Existing datasets for large-scale image diffusion model training, like LAION-5B~\cite{LAION-5b_NEURIPS2022}, lack manual annotations or OCR results for text content. Secondly, as pointed out in~\cite{Liu_CharacterAware_ACL23}, the text encoder used in open-source diffusion models, such as the CLIP text encoder, employs a vocabulary-based tokenizer that has no direct access to the characters, resulting in a diminished sensitivity to individual characters. Lastly, most diffusion models' loss functions are designed to enhance the overall image generation quality and lack of dedicated supervision for the text region.

To address the aforementioned difficulties, we present \textit{AnyText} framework and \textit{AnyWord-3M} dataset. AnyText consists of a text-control diffusion pipeline with two components: auxiliary latent module encodes auxiliary information such as text glyph, position, and masked image into latent space to assist text generation and editing; text embedding module employs an OCR model to encode stroke information as embeddings, which is then fused with image caption embeddings from the tokenizer to render texts blended with the background seamlessly; and finally, a text perceptual loss in image space is introduced to further enhance writing accuracy.

Regarding the functionality, there are five differentiating factors that set apart us from other competitors as outlined in Table~\ref{table:funcs}: a)~\textit{Multi-line}: AnyText can generate text on multiple lines at user-specified positions. b)~\textit{Deformed regions}: it enables writing in horizontally, vertically, and even curved or irregular regions. c)~\textit{Multi-lingual}: our method can generate text in various languages such as Chinese, English, Japanese, Korean, etc. d)~\textit{Text editing}: which provides the capability for modifying the text content within the provided image in consistent font style. e)~\textit{Plug-and-play}: AnyText can be seamlessly integrated with stable diffusion models and empowering them with the ability to generate text. We present some selected examples in Fig.~\ref{fig:samples} and Appendix~\ref{app: more_examples}.

\begin{table}
    \vspace{-3mm}
    \small
    \centering
    \vspace{-0.15in}
    \caption{\small Comparison of AnyText with other competitors based on functionality.}
    \vspace{0.0in}
    \setlength{\tabcolsep}{1.35mm}{
    \begin{tabular}{l|c|c|c|c|c}
    \hline
    Functionality   & Multi-line  & Deformed regions & Multi-lingual & Text editing  & Plug-and-play  \\ \hline
    GlyphDraw      & $\times$ & $\times$ & $\times$  & $\times$ & $\times$ \\
    TextDiffuser   & \checkmark & $\times$ & $\times$ & \checkmark & $\times$ \\
    GlyphControl   & \checkmark & $\times$ & $\times$ & $\times$ & \checkmark  \\
\hline
    AnyText  & \checkmark & \checkmark & \checkmark & \checkmark & \checkmark \\ \hline
    \end{tabular}}
    \vspace{-0.15in}
    \label{table:funcs}
\end{table}
\section{Related Works}
\noindent \textbf{Text-to-Image Synthesis}.~In recent years, significant strides has been made in text-to-image synthesis using denoising diffusion probabilistic models ~\cite{Ho_DDPM_NIPS20, Ramesh_DALLE_icml21, Song_SDE_ICLR21, Dhariwal_GuidedDiffusion_NIPS21, Nichol_ImprovedDDPM_ICML21, Saharia_Imagen_NIPS22, Ramesh_DALLE2_corr22, Rombach_LDM_CVPR22, Chang_Muse_Corr23}. These models have advanced beyond simple image generation and have led to developments in interactive image editing ~\cite{Meng_SDEdit_ICLR22, Gal_Inversion_ICLR23, Brooks_InstructPix_Corr22} and techniques incorporating additional conditions, such as masks and depth maps ~\cite{Rombach_LDM_CVPR22}. Research is also exploring the area of multi-condition controllable synthesis ~\cite{Zhang_ControlNet_Corr23, Mou_T2I_Corr23, Huang_Composer_Corr23}. Compositing subjects into scenes presents more specific challenges, and approaches like ELITE ~\cite{Wei_ELITE_Corr23}, UMM-Diffusion ~\cite{Ma_UnifiedDiffusion_Corr23}, and InstantBooth ~\cite{Shi_InstantBooth_Corr23} utilize the features from CLIP image encoder to encode the visual concept into textual word embeddings. Similarly, DreamIdentity \cite{Chen_DreamIdentity_Corr23} developed a specially designed image encoder to achieve better performance for the word embedding enhancing scheme.

\noindent \textbf{Text Generation}.~Progress in image synthesis has been substantial, but integrating legible text into images remains challenging ~\cite{Rombach_LDM_CVPR22, Saharia_Imagen_NIPS22}. Recent research has focused on three key aspects of text generation:

\textit{Control Condition.} Introducing glyph condition in latent space has been a predominant approach in many recent methods. GlyphDraw ~\cite{Ma_glyphdraw_Corr23} originally used an explicit glyph image as condition, with characters rendered at the center. GlyphControl ~\cite{Yang_GlyphControl_Corr23} further extends it by aligning the text based on its location, which also incorporates font size and text box position in an implicit manner. TextDiffuser~\cite{Chen_TextDiffuser_Corr23} utilizes a character-level segmentation mask as control condition, and in addition, it introduces a masked image to simultaneously learn text generation and text in-painting branches. In our work, we adopt a similar way as GlyphControl to render glyph image but incorporate position and masked image as additional conditions. This design enables AnyText to generate text in curved or irregular regions, and handle text generation and editing simultaneously.

\textit{Text Encoder.} The text encoder plays a crucial role in generating accurate visual text. Recent methods such as Imagen~\cite{Saharia_Imagen_NIPS22}, eDiff-I~\cite{Balaji_eDiffiI_Corr22}, and Deepfloyd IF~\cite{DeepFloyd_23} achieve impressive results by leveraging large-scale language models (e.g., T5-XXL). However, most image generation models still rely on character-blind text encoders, and even character-aware text encoders struggle with non-Latin text generation like Chinese, Japanese, and Korean \cite{Liu_CharacterAware_ACL23}. To address Chinese rendering, GlyphDraw fine-tunes the text encoder on Chinese images and utilizes the CLIP image encoder for glyph embeddings \cite{Ma_glyphdraw_Corr23}. DiffUTE replaces the text encoder with a pre-trained image encoder to extract glyphs in image editing scenarios \cite{Chen_DiffUTE_Corr23}. In AnyText, we propose a novel approach to transform the text encoder by integrating semantic and glyph information. This aims to achieve seamless integration of generated text with the background and enable multi-language text generation.

\textit{Perceptual Supervision.} OCR-VQGAN~\cite{Rodriguez_OCR-VQGAN_WACV23} employs a pre-trained OCR detection model to extract features from images, and supervise text generation by constraining the differences between multiple intermediate layers. In contrast, TextDiffuser~\cite{Chen_TextDiffuser_Corr23} utilizes a character-level segmentation model to supervise the accuracy of each generated character in the latent space. This approach requires a separately trained segmentation model, and the character classes are also limited. In AnyText, we utilize an OCR recognition model that excels in stroke and spelling to supervise the text generation within the designated text region only. This approach provides a more direct and effective form of supervision for ensuring accurate and high-quality text generation.

\section{Methodology}

\begin{figure}[t]
 \centering
 \vspace{-5mm}
 \includegraphics[width=\textwidth]{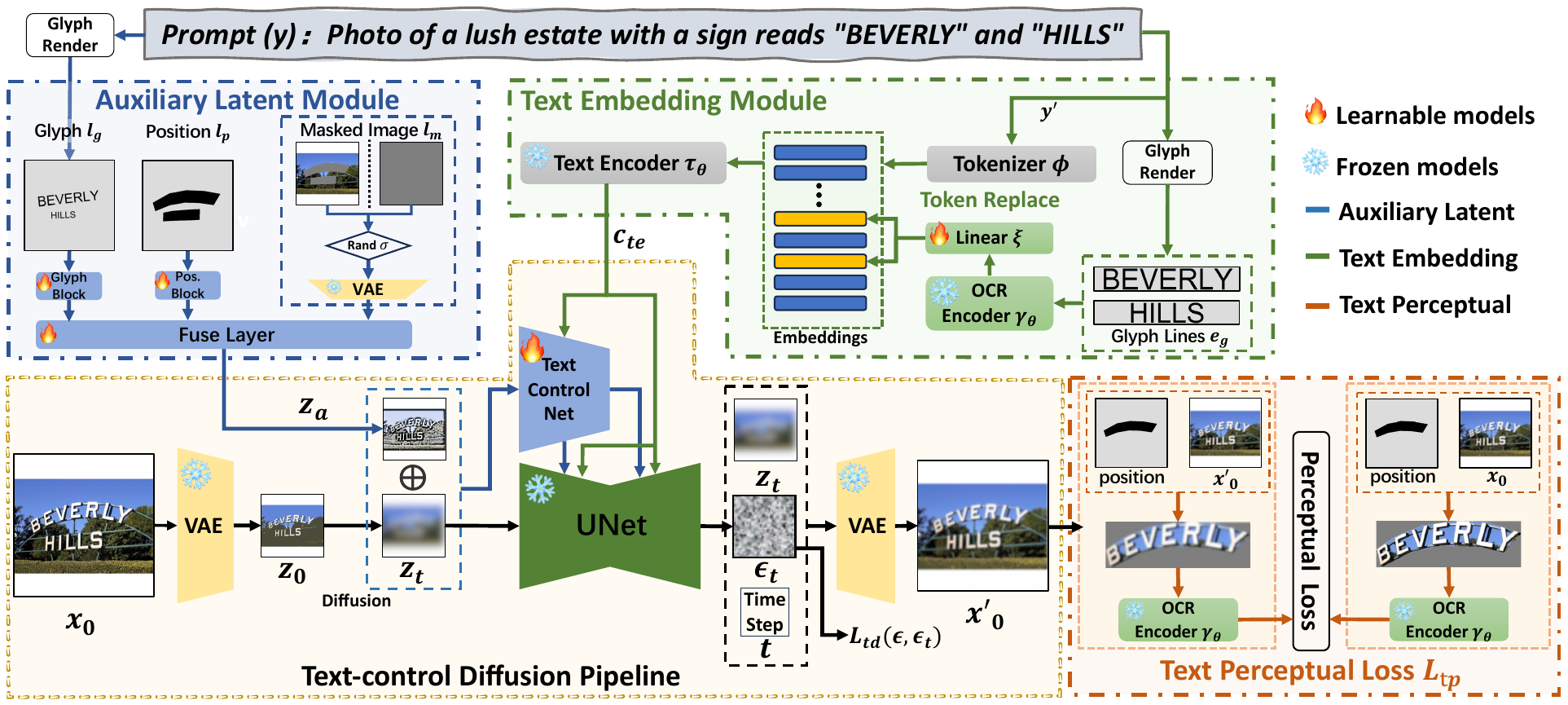}
 \caption{The framework of AnyText, which includes text-control diffusion pipeline, auxiliary latent module, text embedding module, and text perceptual loss.}
 \label{fig:framework}
\end{figure}

As depicted in Fig.~\ref{fig:framework}, the AnyText framework comprises a text-control diffusion pipeline with two primary components (auxiliary latent module and text embedding module). The overall training objective is defined as:
\begin{equation}
\mathcal{L}=\mathcal{L}_{td} + \lambda*\mathcal{L}_{tp}
\label{equ: overall}
\end{equation}
where $\mathcal{L}_{td}$ and $\mathcal{L}_{tp}$ are text-control diffusion loss and text perceptual loss, and $\lambda$ is used to adjust the weight ratio between two loss functions. In the following sections, we will introduce the text-control diffusion pipeline, auxiliary latent module, text embedding module, and text perceptual loss in detail. 

\subsection{Text-control diffusion pipeline}

In the text-control diffusion pipeline, we generate the latent representation $z_0 \in \mathbb{R}^{h \times w \times c}$ by applying Variational Autoencoder (VAE) \cite{KingmaW13_vae_iclr14} \begin{Large}$\varepsilon$\end{Large} on the input image $x_0 \in \mathbb{R}^{H \times W \times 3}$. Here, $h \times w$ represents the feature resolution downsampled by a factor of $f$, and $c$ denotes the latent feature dimension. Then latent diffusion algorithms progressively add noise to the $\textbf{\textit{z}}_0$ and produce a noisy latent image $\textbf{\textit{z}}_t$, where $t$ represents the time step. Given a set of conditions including time step $t$, auxiliary feature $\textbf{\textit{z}}_a\in\mathbb{R}^{h\times w\times c}$ produced by auxiliary latent module, as well as text embedding $\textbf{\textit{c}}_{te}$ produced by text embedding module, text-control diffusion algorithm applies a network $\epsilon_{\theta}$ to predict the noise added to the noisy latent image $\textbf{\textit{z}}_t$ with objective:
\begin{equation}
\mathcal{L}_{td}=\mathbb{E}_{\textbf{\textit{z}}_0,\textbf{\textit{z}}_a, \textbf{\textit{c}}_{te}, \textbf{\textit{t}}, \epsilon\sim\mathcal{N}(0,1)}\left[\|\epsilon-\epsilon_{\theta}(\textbf{\textit{z}}_t, \textbf{\textit{z}}_a, \textbf{\textit{c}}_{te}, \textbf{\textit{t}})\|_{2}^{2}\right]
\end{equation}
\noindent where $\mathcal{L}_{td}$ is the text-control diffusion loss. More specifically, to control the generation of text, we add $z_a$ with $z_t$ and fed them into a trainable copy of UNet's encoding layers referred to as TextControlNet, and fed $z_t$ into a parameter-frozen UNet. This enables TextControlNet to focus on text generation while preserving the base model's ability to generate images without text. Moreover, through modular binding, a wide range of base models can also generate text. Please refer to Appendix~\ref{app: flexibility} for more details.

\subsection{Auxiliary latent module}
\label{subsec:latent_aux}
\begin{figure}[htbp]
 \centering
 \includegraphics[width=\textwidth]{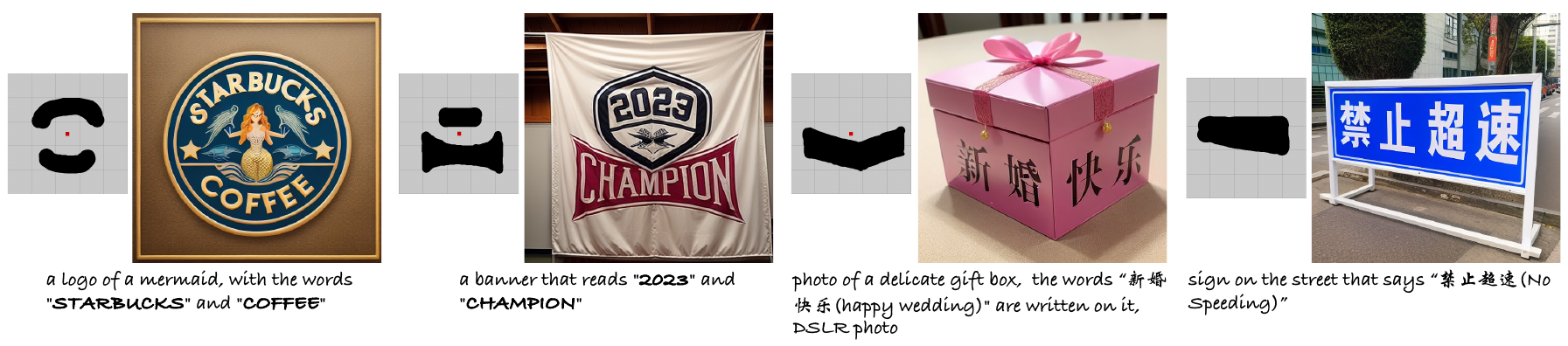}
 \caption{Illustration of generating text in curved or irregular regions, left images are text positions provided by the user.}
 \label{fig:curve}
\end{figure}

In AnyText, three types of auxiliary conditions are utilized to produce latent feature map $\textbf{\textit{z}}_a$: glyph $l_g$, position $l_p$, and masked image $l_m$. Glyph $l_g$ is generated by rendering texts using a uniform font style (i.e., ``Arial Unicode") onto an image based on their locations. Accurately rendering characters in curved or irregular regions is considerably challenging. Therefore, we simplify the process by rendering characters based on the enclosing rectangle of the text position. By incorporating specialized position $l_p$, we can still generate text in non-rectangular regions, as illustrated in Fig.~\ref{fig:curve}. Position $l_p$ is generated by marking text positions on an image. In the training phase, the text positions are obtained either from OCR detection or through manual annotation. In the inference phase, $l_p$ is obtained from the user's input, where they specify the desired regions for text generation. Moreover, the position information allows the text perceptual loss to precisely target the text area. Details regarding this will be discussed in Sec.~\ref{subsec:perceptual_loss}.
The last auxiliary information is masked image $l_m$, which indicates what area in image should be preserved during the diffusion process. In the text-to-image mode, $l_m$ is set to be fully masked, whereas in the text editing mode, $l_m$ is set to mask the text regions. During training, the text editing mode ratio is randomly switched with a probability of $\sigma$.

To incorporate the image-based conditions, we use glyph block and position block to downsample glyph $l_g$ and position $l_p$, and VAE encoder \begin{Large}$\varepsilon$\end{Large} to downsample the masked image $l_m$, respectively. The glyph block $G$ and position block $P$ both contain several stacked convolutional layers. After transforming these image-based conditions into feature maps that match the spatial size of $z_t$, we utilize a convolutional fusion layer $f$ to merge $l_g$, $l_p$, and $l_m$, resulting in a generated feature map denoted as $z_a$, which can be represented as:
\begin{equation}
\textbf{\textit{z}}_a  =  f(G(l_g) + P(l_p) + \begin{Large}{\varepsilon}\end{Large}(l_m))
\end{equation}
\noindent where $\textbf{\textit{z}}_a$ shares the same number of channels as $\textbf{\textit{z}}_t$.

\subsection{Text embedding module}
\label{subsec:text_embedding}
\begin{figure}[htbp]
 \centering
 \includegraphics[width=\textwidth]{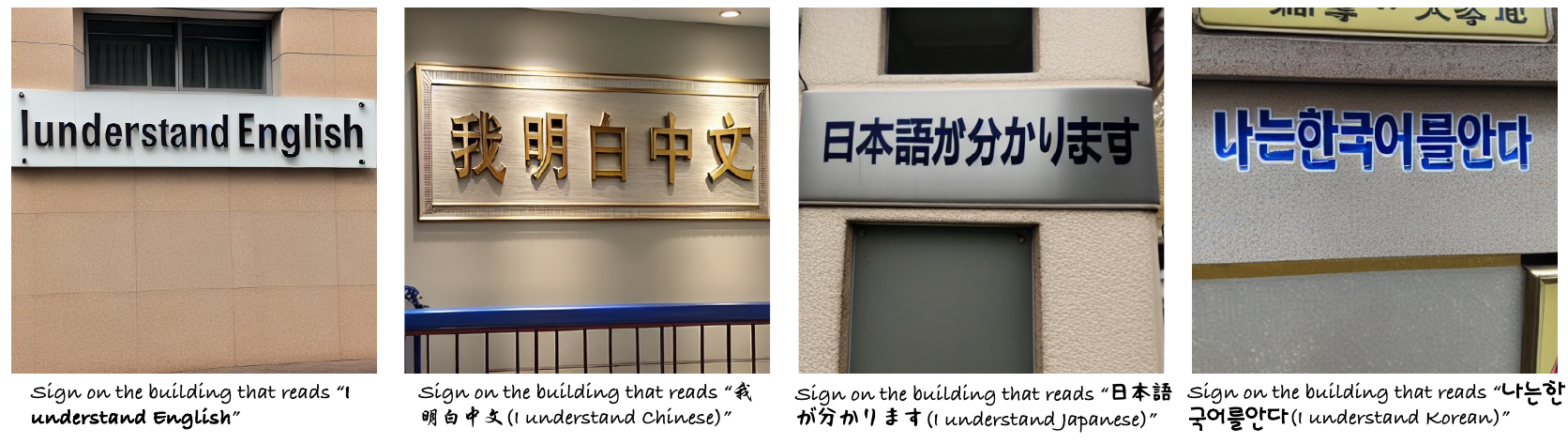}
 \caption{Illustration of generating text in multiple languages.}
 \label{fig:multilingual}
\end{figure}

Text encoders excel at extracting semantic information from caption, but semantic information of the text to be rendered is negligible. Additionally, most pre-trained text encoders are trained on Latin-based data and can't understand other languages well. In AnyText, we propose a novel approach to address the problem of multilingual text generation. Specifically, we render glyph lines into images, encode glyph information, and replace their embeddings from caption tokens. The text embeddings are not learned character by character, but rather utilize a pre-trained visual model, specifically the recognition model of PP-OCRv3~\cite{chen_ppocrv3_corr22}. The replaced embeddings are then fed into a transformer-based text encoder as tokens to get fused intermediate representation, which will then be mapped to the intermediate layers of the UNet using a cross-attention mechanism. Due to the utilization of image rendering for text instead of relying solely on language-specific text encoders, our approach significantly enhances the generation of multilingual text, as depicted in Fig.~\ref{fig:multilingual}.

Next, we provide a detailed explanation of the text embedding module. The representation $\textbf{\textit{c}}_{te}$ that combines both text glyph and caption semantic information is defined as
\begin{equation}
\textbf{\textit{c}}_{te}=\tau_{\theta}(\phi({y'}), \xi(\gamma_{\theta}(e_g)))
\end{equation}
\noindent where $y'$ is the processed input caption $y$ that each text line to be generated (enclosed within double quotation marks) is replaced with a special placeholder $S_*$. Then after tokenization and embedding lookup denoted as $\phi(\cdot)$, the caption embeddings are obtained. Then, each text line is rendered onto an image, denoted as $e_g$. Note that $e_g$ is generated by only rendering a single text line onto an image in the center, while $l_g$ in Sec.~\ref{subsec:latent_aux} is produced by rendering all text lines onto a single image on their locations. The image $e_g$ is then fed into an OCR recognition model $\gamma_{\theta}$ to extract the feature before the last fully connected layer as text embedding, then a linear transformation $\xi$ is applied to ensure its size matches caption embeddings, and replace it with embedding of $S_*$. Finally, all token embeddings are encoded using a CLIP text encoder $\tau_{\theta}$.

\subsection{Text perceptual loss}
\label{subsec:perceptual_loss}
We propose a text perceptual loss to further improve the accuracy of text generation. Assuming $\mathcal{\varepsilon}_t$ represents the noise predicted by the denoiser network $\epsilon_{\theta}$, we can combine the time step $t$ and noisy latent image $z_t$ to predict $z_0$ as described in ~\cite{Ho_DDPM_NIPS20}. This can be further used with the VAE decoder to obtain an approximate reconstruction of the original input image, denoted as $x'_{0}$. By transitioning from latent space to image space, we can further supervise text generation at a pixel-wise level. With the help of the position condition $l_p$, we can accurately locate the region of generated text. We aim to compare this region with the corresponding area in the original image $x_0$,  and focus solely on the writing correctness of the text itself, excluding factors such as background, deviations in character positions, colors, or font styles. Thus, we employ the PP-OCRv3 model, as mentioned in Sec.~\ref{subsec:text_embedding}, as the image encoder. 
By processing $x_0$ and $x'_0$ at position $p$ through operations such as cropping, affine transformation, padding, and normalization, we obtain the images $p_g$ and $p'_g$ to be used as inputs for the OCR model. We utilize the feature maps $\hat{m}_{p},\hat{m'}_{p}\in\mathbb{R}^{h\times w\times c}$ before the fully connected layer to represent the textual writing information in the original and predicted image at position $p$, respectively. The text perceptual loss is expressed as
\begin{equation}
\mathcal{L}_{tp}=\sum_p\frac{\varphi(t)}{hw}\sum_{h,w}\|\hat{m}_{p}-\hat{m'}_{p}\|_{2}^{2}
\end{equation}
By imposing a Mean Squared Error (MSE) penalty, we attempt to minimize the discrepancies between the predicted and original image in all text regions. As time step $t$ is related to the text quality in predicted image $x'_0$, we need to design a weight adjustment function $\varphi(t)$. It has been found that setting $\varphi(t) = \bar{\alpha_t}$ is a good choice, where $\bar{\alpha_t}$ is the coefficient of diffusion process introduced in ~\cite{Ho_DDPM_NIPS20}.

\section{Dataset and Benchmark}
\label{sec:dataset}
Currently, there is a lack of publicly available datasets specifically tailored for text generation tasks, especially those involving non-Latin languages. Therefore, we propose \textit{AnyWord-3M},  a large-scale multilingual dataset from publicly available images. The sources of these images include  Noah-Wukong~\cite{wukong_corr22}, LAION-400M~\cite{laion_400m_corr21}, as well as datasets used for OCR recognition tasks such as ~\cite{ArT}, ~\cite{COCO-Text}, ~\cite{RCTW},~\cite{LSVT}, ~\cite{MLT}, ~\cite{MTWI}, ~\cite{ReCTS}. These images cover a diverse range of scenes that contain text, including street views, book covers, advertisements, posters, movie frames, etc. Except for the OCR datasets, where the annotated information is used directly, all other images are processed using the PP-OCRv3~\cite{chen_ppocrv3_corr22} detection and recognition models. Then, captions are regenerated using BLIP-2~\cite{Li_BLIP2_Corr23}. Please refer to Appendix~\ref{app: dataset} for more details about dataset preparation.

Through strict filtering rules and meticulous post-processing, we obtained a total of 3,034,486 images, with over 9 million lines of text and more than 20 million characters or Latin words. We randomly extracted 1000 images from both Wukong and LAION subsets to create the evaluation set called AnyText-benchmark. These two evaluation sets are specifically used to evaluate the accuracy and quality of Chinese and English generation, respectively. The remaining images are used as the training set called AnyWord-3M, among which approximately 1.6 million are in Chinese, 1.39 million are in English, and 10k images in other languages, including Japanese, Korean, Arabic, Bangla, and Hindi. For a detailed statistical analysis and randomly selected example images, please refer to Appendix~\ref{app: stastic}.

For the AnyText-benchmark, we utilized three evaluation metrics to assess the accuracy and quality of text generation. Firstly, we employed the Sentence Accuracy (Sen. Acc) metric, where each generated text line was cropped according to the specified position and fed into the OCR model to obtain predicted results. Only when the predicted text completely matched the ground truth was it considered correct. Additionally, we employed another less stringent metric, the Normalized Edit Distance (NED) to measure the similarity between two strings. As we used PP-OCRv3 for feature extraction during training, to ensure fairness, we chose another open-source model called DuGuangOCR~\cite{DuGuangOCR} for evaluation, which also performs excellently in both Chinese and English recognition. However, relying solely on OCR cannot fully capture the image quality. Therefore, we introduced the Frechet Inception Distance (FID) to assess the distribution discrepancy between generated images and real-world images.
\section{Experiments}

\subsection{Implementation Details}
\label{subsec: imp_details}
Our training framework is implemented based on ControlNet\footnote{https://github.com/lllyasviel/ControlNet}, and the model's weights are initialized from SD1.5\footnote{https://huggingface.co/runwayml/stable-diffusion-v1-5}. In comparison to ControlNet, we only increased the parameter size by 0.34\% and the inference time by 1.04\%, refer to ~\ref{app: param_size} for more details. Our model was trained on the AnyWord-3M dataset for 10 epochs using 8 Tesla A100 GPUs. We employed a progressive fine-tuning strategy, where the editing branch was turned off for the first 5 epochs, then activated with a probability of $\sigma=0.5$ for the next 3 epochs. In the last 2 epochs, we enabled the perceptual loss with a weight coefficient of $\lambda=0.01$. Image dimensions of $l_g$ and $l_p$ are set to be 1024x1024 and 512x512, while $e_g$, $p_g$, and $p'_g$ are all set to be 80x512. We use AdamW optimizer with a learning rate of 2e-5 and a batch size of 48. During sampling process, based on the statistical information from ~\ref{app: stastic}, a maximum of 5 text lines from each image and 20 characters from each text line were chosen to render onto the image, as this setting can cover the majority of cases in the dataset.

Recently, we found that the quality of OCR annotations in the training data has a significant impact on the text generation metrics. We made minor improvements to the method and dataset, then fine-tuned the original model, resulting in the AnyText-v1.1, which achieved significant performance improvements. Details of the updated model are provided in Appendix~\ref{app: details_v1.1}. It should be noted that all example images used in this article are still generated by the v1.0 model.

\subsection{Comparison results}

\subsubsection{Quantitative Results}
We evaluated existing competing methods, including ControlNet~\cite{Zhang_ControlNet_Corr23}, TextDiffuser~\cite{Chen_TextDiffuser_Corr23}, and GlyphControl~\cite{Yang_GlyphControl_Corr23}, using the benchmark and metrics mentioned in Sec.~\ref{sec:dataset}. To ensure fair evaluation, all methods employed the DDIM sampler with 20 steps of sampling, a CFG-scale of 9, a fixed random seed of 100, a batch size of 4, and the same positive and negative prompt words. The quantitative comparison results can be found in Table~\ref{table:quanti_res}. Additionally, we provide some generated images from AnyText-benchmark in Appendix~\ref{app:eval-examples}.

\begin{table}
    \vspace{-8mm}
    \small
    \centering
    \caption{\small Quantitative comparison of AnyText and competing methods. \dag is trained on LAION-Glyph-10M, and \ddag is fine-tuned on TextCaps-5k. All competing methods are evaluated using their officially released models. }
    \vspace{0.0in}
    \setlength{\tabcolsep}{1.5mm}{
    \begin{tabular}{l|c|c|c|c|c|c}
    \hline
    \multirow{2}{*}{Methods}   & \multicolumn{3}{c|}{English}  & \multicolumn{3}{c}{Chinese}  \\
    \cline{2-7}  & Sen. ACC↑ & NED↑ & FID↓ & Sen. ACC↑ & NED↑ & FID↓  \\ \hline
    ControlNet   & 0.5837 & 0.8015 & 45.41 & 0.3620 & 0.6227 & 41.86  \\
    TextDiffuser  & 0.5921 & 0.7951 & 41.31 & 0.0605 & 0.1262 & 53.37  \\
    GlyphControl\dag & 0.3710 & 0.6680 & 37.84 & 0.0327 & 0.0845 & 34.36 \\
    GlyphControl\ddag & 0.5262 & 0.7529 & 43.10 & 0.0454 & 0.1017 & 49.51 \\
    \hline
    AnyText\textbf{-v1.0} & 0.6588 & 0.8568 & 35.87 & 0.6634 & 0.8264 & \textbf{28.46} \\
    AnyText\textbf{-v1.1} & \textbf{0.7239} & \textbf{0.8760} & \textbf{33.54} & \textbf{0.6923} & \textbf{0.8396} & 31.58 \\
    \hline
\end{tabular}}
    \vspace{-0.15in}
    \label{table:quanti_res}
\end{table}

From the results, we can observe that AnyText outperforms competing methods in both Chinese and English text generation by a large margin, in terms of OCR accuracy (Sen.ACC, NED) and realism (FID). It is worth mentioning that our training set only consists of 1.39 million English data, while TextDiffuser and GlyphControl are trained on 10 million pure English data. An interesting phenomenon is that ControlNet(with canny control) tends to randomly generate pseudo-text in the background, evaluation methods as in ~\cite{Chen_TextDiffuser_Corr23} that utilize OCR detection and recognition models may yield low evaluation scores. However, in our metric, we focus only on the specified text generation areas, and we find that ControlNet performs well in these areas. Nevertheless, the style of the generated text from ControlNet appears rigid and monotonous, as if it was pasted onto the background, resulting in a poor FID score. Regarding Chinese text generation, both TextDiffuser and GlyphControl can only generate some Latin characters, punctuation marks or numbers within a Chinese text. In AnyText, our Chinese text generation accuracy surpasses all other methods. In the stringent evaluation metric of Sen. ACC, we achieve an accuracy of over 66\%. Additionally, AnyText obtains the lowest FID score, indicating superior realism in the generated text.

\subsubsection{Qualitative Results}
\begin{figure}[htbp]
 \centering
 \vspace{-2mm}
 \includegraphics[width=0.86\textwidth]{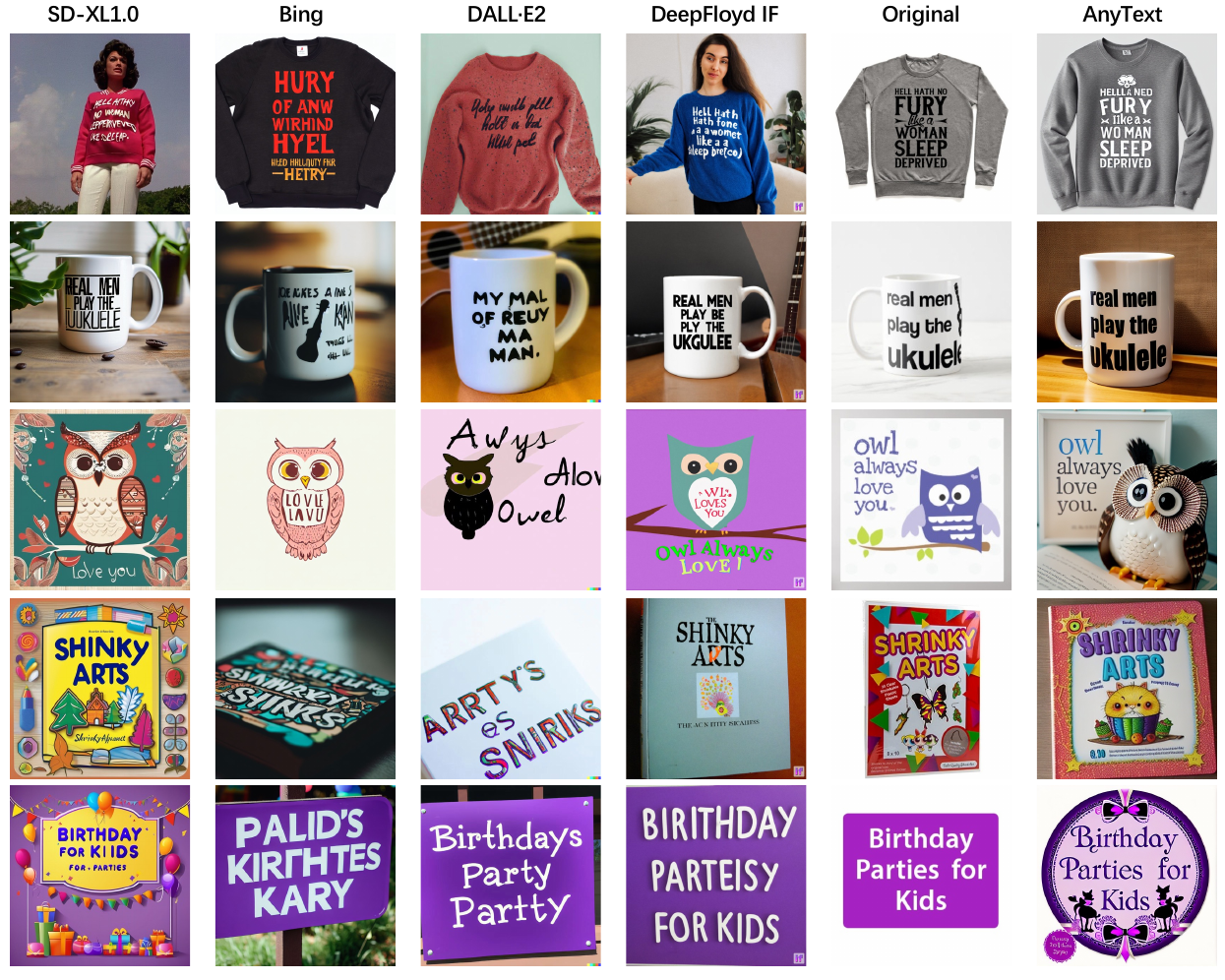}
 \caption{Qualitative comparison of AnyText and state-of-the-art models or APIs in English text generation. All captions are selected from the English evaluation dataset in AnyText-benchmark.}
 \vspace{-5mm}
 \label{fig:compare_sota}
\end{figure}

\begin{figure}[htbp]
 \centering
 \vspace{-2mm}
 \includegraphics[width=0.86\textwidth]{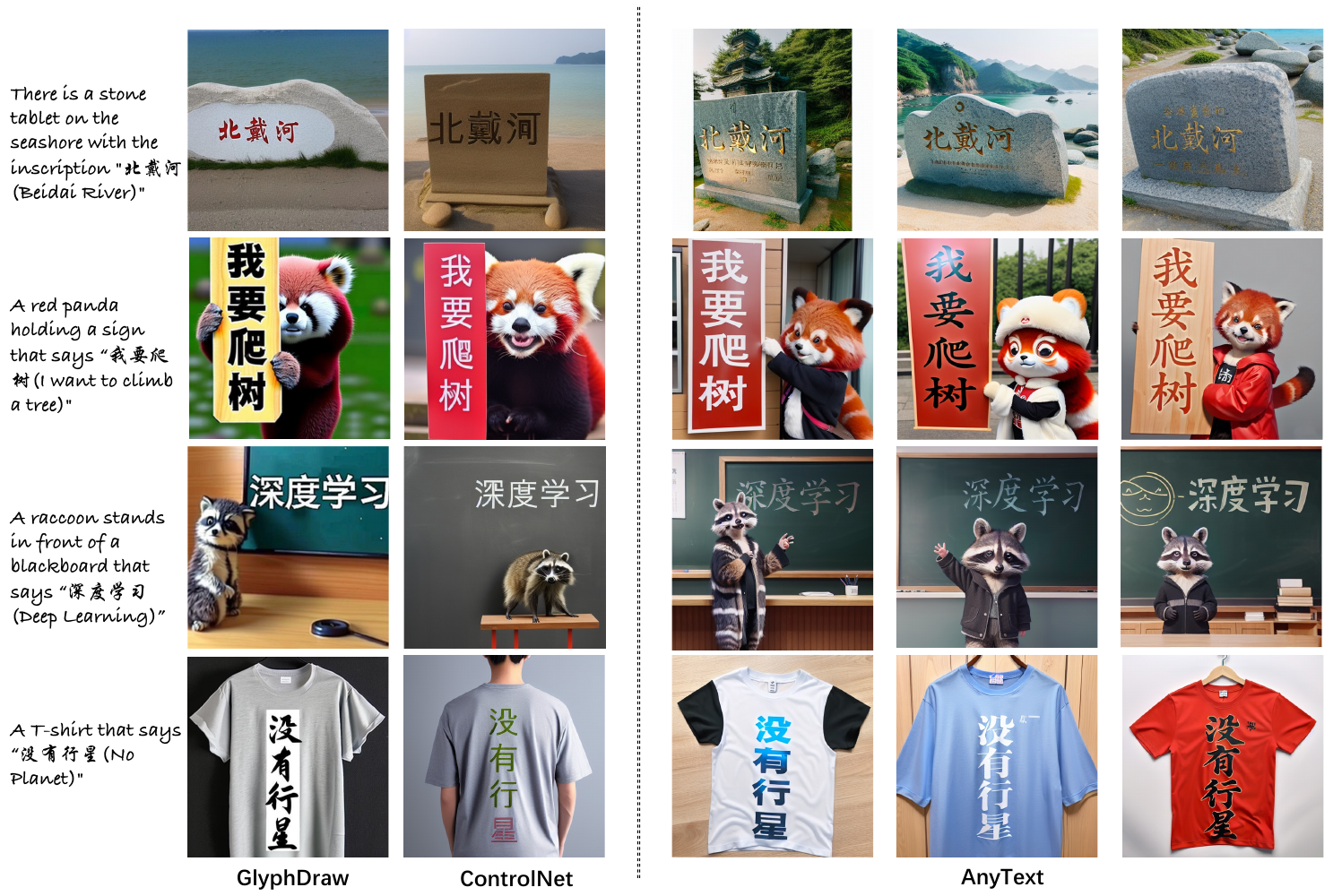}
 \caption{Comparative examples between GlyphDraw, ControlNet, and AnyText in Chinese text generation, all taken from the original GlyphDraw paper.}
 \vspace{-5mm}
 \label{fig:compareGlyphDraw}
\end{figure}

Regarding the generation of the English text, we compared our model with state-of-the-art models or APIs in the field of text-to-image generation, such as SD-XL1.0, Bing Image Creator\footnote{https://www.bing.com/create}, DALL-E2, and DeepFloyd IF, as shown in Fig.~\ref{fig:compare_sota}. These models have shown significant improvements in text generation compared to previous works . However, there is still a considerable gap between them and professional text generation models. Regarding the generation of Chinese text, the complexity of strokes and the vast number of character categories pose significant challenges. GlyphDraw~\cite{Ma_glyphdraw_Corr23} is the first method that addresses this task. Due to the lack of open-source models or APIs, we were unable to conduct quantitative evaluation and instead relied on qualitative comparisons using examples from GlyphDraw paper. Meanwhile, since ControlNet can also generate Chinese text effectively, we included it as well. As shown in Fig.~\ref{fig:compareGlyphDraw}, AnyText shows superior integration of generated text with background, such as text carved into stone, reflections on signboards with words, chalk-style text on a blackboard, and slightly distorted text influenced by clothing folds.

\subsection{Ablation study}
\label{subsec: ablation}
In this part, we extracted 200k images (with 160k in Chinese) from AnyWord-3M as the training set, and used the Chinese evaluation dataset in AnyText-benchmark to validate the effectiveness of each submodule in AnyText. Each model was trained for 15 epochs, requiring 60 hours on 8 Tesla V100 GPUs. The training parameters were kept consistent with those mentioned in Sec.~\ref{subsec: imp_details}. By analyzing the experimental results in Table~\ref{table:ablation}, we draw the following conclusions: 

\textit{Editing}: Comparing Exp.1 and Exp.2 we observed a slight decrease when editing branch is enabled. This is because text generation and editing are two different tasks, and enabling editing branch increases the difficulty of model convergence. To focus on analyzing the text generation part, we disabled editing in all subsequent experiments by setting the probability parameter $\sigma$ to 0.

\textit{Text Embedding}: Exp.$2\sim5$ validate the effectiveness of the text embedding module. In Exp.2, captions are encoded directly by the CLIP text encoder. Exp.3 employs the CLIP vision encoder(vit) as the feature extractor, but yielded unsatisfactory results, likely due to images with only rendered text falling into the Out-of-Distribution data category for the pre-trained CLIP vision model, hindering its text stroke encoding capabilities. Exp.4 experimented with a trainable module composed of stacked convolutions and an average pooling layer(conv), which struggled without tailored supervision. In Exp.5, utilizing the pre-trained OCR model(PP-OCRv3) resulted in a significant 25.7\% increase in Sen. Acc metric compared to Exp.2.

\textit{Position}: Comparing Exp.5 and Exp.6, although the rendered text in $l_g$ contains positional information implicitly, the inclusion of a more accurate position $l_p$ further improves performance and enables the model to generate text in irregular regions.

\textit{Text Perceptual Loss}: Exp.$7\sim9$ validate the effectiveness of text perceptual loss. Upon conducting three experiments, we found that $\lambda$ = 0.01 yielded the best results, with a 4.0\% improvement compared to Exp.5 in Sen. Acc metric. It is worth noting that perceptual loss necessitates transitioning from latent space to image space, which slows down the training process. Therefore, for Exp.$7\sim9$, we started training from epoch 13 of Exp.5 and trained only for the remaining 2 epochs. Despite this, perceptual loss demonstrated significant improvements.

\begin{table}
    \small
    \centering
    \caption{\small Ablation experiments of AnyText on a small-scale dataset from AnyWord-3M. The results validate the effectiveness of each submodule in AnyText.}
    \vspace{0.0in}
    \setlength{\tabcolsep}{1.35mm}{
    \begin{tabular}{c|c|c|c|c|c|c|c}
    \hline
    \multirow{2}{*}{Exp. \textnumero}   & \multirow{2}{*}{Editing} & \multirow{2}{*}{Position} & \multirow{2}{*}{Text Embedding} & \multirow{2}{*}{Perceptual Loss} & \multirow{2}{*}{$\lambda$} & \multicolumn{2}{c}{Chinese} \\
    \cline{7-8} & & & & & & Sen. ACC↑ & NED↑  \\ \hline
    1    & \checkmark & \checkmark & $\times$ & $\times$ & - & 0.1552 & 0.4070 \\
    2    & $\times$ & \checkmark & $\times$ & $\times$ & - & 0.2024 & 0.4649 \\
    3    & $\times$ & \checkmark &  vit  & $\times$ & - & 0.1416 & 0.3809 \\
    4    & $\times$ & \checkmark &  conv & $\times$ & - & 0.1864 & 0.4402 \\
    5    & $\times$ & \checkmark &  ocr  & $\times$ & - & 0.4595 & 0.7072 \\
    6    & $\times$ & $\times$ & ocr & $\times$ & - & 0.4472 & 0.6974 \\
    7    & $\times$ & \checkmark & ocr & \checkmark & 0.003 & 0.4848 & 0.7353 \\
    8    & $\times$ & \checkmark & ocr & \checkmark & 0.01 & \textbf{0.4996} & \textbf{0.7457} \\
    9    & $\times$ & \checkmark & ocr & \checkmark & 0.03 & 0.4659 & 0.7286 \\ \hline
    \end{tabular}}
    \label{table:ablation}
\end{table}

\section{Conclusion and Limitations}
In this paper, we delve into the extensively researched problem of text generation in the field of text-to-image synthesis. To address this challenge, we propose a novel approach called AnyText, which is a diffusion-based multi-lingual text generation and editing framework. Our approach incorporates an auxiliary latent module that combines text glyph, position, and masked image into a latent space. Furthermore, in the text embedding module, we leverage an OCR model to extract glyph information and merge it with the semantic details of the image caption, thereby enhancing the consistency between the text and the background. To improve writing accuracy, we employ text-control diffusion loss and text perceptual loss during training. In terms of training data, we present the AnyWord-3M dataset, comprising 3 million text-image pairs in multiple languages with OCR annotations. To demonstrate the superior performance of AnyText, we conduct comprehensive experiments on the proposed AnyText-benchmark, showcasing its superiority over existing methods. Moving forward, our future work will focus on exploring the generation of extremely small fonts and investigating text generation with controllable attributes.

\bibliography{iclr2024_conference}

\begin{thebibliography}{43}
\providecommand{\natexlab}[1]{#1}
\providecommand{\url}[1]{\texttt{#1}}
\expandafter\ifx\csname urlstyle\endcsname\relax
  \providecommand{\doi}[1]{doi: #1}\else
  \providecommand{\doi}{doi: \begingroup \urlstyle{rm}\Url}\fi

\bibitem[ArT()]{ArT}
ArT.
\newblock Icdar2019 robust reading challenge on arbitrary-shaped text.
\newblock https://rrc.cvc.uab.es/?ch=14, 2019.

\bibitem[Balaji et~al.(2022)Balaji, Nah, Huang, Vahdat, Song, Kreis, Aittala, Aila, Laine, Catanzaro, Karras, and Liu]{Balaji_eDiffiI_Corr22}
Yogesh Balaji, Seungjun Nah, Xun Huang, Arash Vahdat, Jiaming Song, Karsten Kreis, Miika Aittala, Timo Aila, Samuli Laine, Bryan Catanzaro, Tero Karras, and Ming{-}Yu Liu.
\newblock ediff-i: Text-to-image diffusion models with an ensemble of expert denoisers.
\newblock \emph{arXiv preprint}, 2022.

\bibitem[Brooks et~al.(2022)Brooks, Holynski, and Efros]{Brooks_InstructPix_Corr22}
Tim Brooks, Aleksander Holynski, and Alexei~A. Efros.
\newblock Instructpix2pix: Learning to follow image editing instructions.
\newblock \emph{arXiv preprint}, abs/2211.09800, 2022.

\bibitem[Chang et~al.(2023)Chang, Zhang, Barber, Maschinot, Lezama, Jiang, Yang, Murphy, Freeman, Rubinstein, Li, and Krishnan]{Chang_Muse_Corr23}
Huiwen Chang, Han Zhang, Jarred Barber, Aaron Maschinot, Jos{\'{e}} Lezama, Lu~Jiang, Ming{-}Hsuan Yang, Kevin Murphy, William~T. Freeman, Michael Rubinstein, Yuanzhen Li, and Dilip Krishnan.
\newblock Muse: Text-to-image generation via masked generative transformers.
\newblock \emph{arXiv preprint}, abs/2301.00704, 2023.

\bibitem[Chen et~al.(2023{\natexlab{a}})Chen, Xu, Gu, Lan, Zheng, Li, Meng, Zhu, and Wang]{Chen_DiffUTE_Corr23}
Haoxing Chen, Zhuoer Xu, Zhangxuan Gu, Jun Lan, Xing Zheng, Yaohui Li, Changhua Meng, Huijia Zhu, and Weiqiang Wang.
\newblock Diffute: Universal text editing diffusion model.
\newblock \emph{arXiv preprint}, abs/2305.10825, 2023{\natexlab{a}}.

\bibitem[Chen et~al.(2023{\natexlab{b}})Chen, Huang, Lv, Cui, Chen, and Wei]{Chen_TextDiffuser_Corr23}
Jingye Chen, Yupan Huang, Tengchao Lv, Lei Cui, Qifeng Chen, and Furu Wei.
\newblock Textdiffuser: Diffusion models as text painters.
\newblock \emph{arXiv preprint}, abs/2305.10855, 2023{\natexlab{b}}.

\bibitem[Chen et~al.(2023{\natexlab{c}})Chen, Fang, Liu, He, Huang, Zhang, and Mao]{Chen_DreamIdentity_Corr23}
Zhuowei Chen, Shancheng Fang, Wei Liu, Qian He, Mengqi Huang, Yongdong Zhang, and Zhendong Mao.
\newblock Dreamidentity: Improved editability for efficient face-identity preserved image generation.
\newblock \emph{arXiv preprint}, abs/2307.00300, 2023{\natexlab{c}}.

\bibitem[COCO-Text()]{COCO-Text}
COCO-Text.
\newblock A large-scale scene text dataset based on mscoco.
\newblock https://bgshih.github.io/cocotext, 2016.

\bibitem[DeepFloyd-Lab(2023)]{DeepFloyd_23}
DeepFloyd-Lab.
\newblock Deepfloyd if.
\newblock \emph{https://github.com/deep-floyd/IF}, 2023.

\bibitem[Dhariwal \& Nichol(2021)Dhariwal and Nichol]{Dhariwal_GuidedDiffusion_NIPS21}
Prafulla Dhariwal and Alexander~Quinn Nichol.
\newblock Diffusion models beat gans on image synthesis.
\newblock In \emph{NeurIPS}, pp.\  8780--8794, 2021.

\bibitem[Gal et~al.(2023)Gal, Alaluf, Atzmon, Patashnik, Bermano, Chechik, and Cohen{-}Or]{Gal_Inversion_ICLR23}
Rinon Gal, Yuval Alaluf, Yuval Atzmon, Or~Patashnik, Amit~Haim Bermano, Gal Chechik, and Daniel Cohen{-}Or.
\newblock An image is worth one word: Personalizing text-to-image generation using textual inversion.
\newblock In \emph{ICLR}, 2023.

\bibitem[Gu et~al.(2022)Gu, Meng, Lu, Hou, Niu, Xu, Liang, Zhang, Jiang, and Xu]{wukong_corr22}
Jiaxi Gu, Xiaojun Meng, Guansong Lu, Lu~Hou, Minzhe Niu, Hang Xu, Xiaodan Liang, Wei Zhang, Xin Jiang, and Chunjing Xu.
\newblock Wukong: 100 million large-scale chinese cross-modal pre-training dataset and {A} foundation framework.
\newblock \emph{CoRR}, abs/2202.06767, 2022.

\bibitem[Ho et~al.(2020)Ho, Jain, and Abbeel]{Ho_DDPM_NIPS20}
Jonathan Ho, Ajay Jain, and Pieter Abbeel.
\newblock Denoising diffusion probabilistic models.
\newblock In \emph{NeurIPS}, 2020.

\bibitem[Huang et~al.(2023)Huang, Chen, Liu, Shen, Zhao, and Zhou]{Huang_Composer_Corr23}
Lianghua Huang, Di~Chen, Yu~Liu, Yujun Shen, Deli Zhao, and Jingren Zhou.
\newblock Composer: Creative and controllable image synthesis with composable conditions.
\newblock \emph{arXiv preprint}, abs/2302.09778, 2023.

\bibitem[Inc.(2022)]{midjourney}
Midjourney Inc.
\newblock Midjourney.
\newblock \emph{https://www.midjourney.com/}, 2022.

\bibitem[Kingma \& Welling(2014)Kingma and Welling]{KingmaW13_vae_iclr14}
Diederik~P. Kingma and Max Welling.
\newblock Auto-encoding variational bayes.
\newblock In \emph{2nd International Conference on Learning Representations, {ICLR} 2014, Banff, AB, Canada, April 14-16, 2014, Conference Track Proceedings}, 2014.

\bibitem[Li et~al.(2022)Li, Liu, Guo, Yin, Jiang, Du, Du, Zhu, Lai, Hu, Yu, and Ma]{chen_ppocrv3_corr22}
Chenxia Li, Weiwei Liu, Ruoyu Guo, Xiaoting Yin, Kaitao Jiang, Yongkun Du, Yuning Du, Lingfeng Zhu, Baohua Lai, Xiaoguang Hu, Dianhai Yu, and Yanjun Ma.
\newblock Pp-ocrv3: More attempts for the improvement of ultra lightweight {OCR} system.
\newblock \emph{CoRR}, abs/2206.03001, 2022.

\bibitem[Li et~al.(2023)Li, Li, Savarese, and Hoi]{Li_BLIP2_Corr23}
Junnan Li, Dongxu Li, Silvio Savarese, and Steven C.~H. Hoi.
\newblock {BLIP-2:} bootstrapping language-image pre-training with frozen image encoders and large language models.
\newblock \emph{arXiv preprint}, abs/2301.12597, 2023.

\bibitem[Liu et~al.(2023)Liu, Garrette, Saharia, Chan, Roberts, Narang, Blok, Mical, Norouzi, and Constant]{Liu_CharacterAware_ACL23}
Rosanne Liu, Dan Garrette, Chitwan Saharia, William Chan, Adam Roberts, Sharan Narang, Irina Blok, RJ~Mical, Mohammad Norouzi, and Noah Constant.
\newblock Character-aware models improve visual text rendering.
\newblock In \emph{ACL}, pp.\  16270--16297, 2023.

\bibitem[LSVT()]{LSVT}
LSVT.
\newblock Icdar2019 robust reading challenge on large-scale street view text with partial labeling.
\newblock https://rrc.cvc.uab.es/?ch=16, 2019.

\bibitem[Ma et~al.(2023{\natexlab{a}})Ma, Zhao, Chen, Wang, Niu, Lu, and Lin]{Ma_glyphdraw_Corr23}
Jian Ma, Mingjun Zhao, Chen Chen, Ruichen Wang, Di~Niu, Haonan Lu, and Xiaodong Lin.
\newblock Glyphdraw: Learning to draw chinese characters in image synthesis models coherently.
\newblock \emph{arXiv preprint}, abs/2303.17870, 2023{\natexlab{a}}.

\bibitem[Ma et~al.(2023{\natexlab{b}})Ma, Yang, Wang, Fu, and Liu]{Ma_UnifiedDiffusion_Corr23}
Yiyang Ma, Huan Yang, Wenjing Wang, Jianlong Fu, and Jiaying Liu.
\newblock Unified multi-modal latent diffusion for joint subject and text conditional image generation.
\newblock \emph{arXiv preprint}, abs/2303.09319, 2023{\natexlab{b}}.

\bibitem[Meng et~al.(2022)Meng, He, Song, Song, Wu, Zhu, and Ermon]{Meng_SDEdit_ICLR22}
Chenlin Meng, Yutong He, Yang Song, Jiaming Song, Jiajun Wu, Jun{-}Yan Zhu, and Stefano Ermon.
\newblock Sdedit: Guided image synthesis and editing with stochastic differential equations.
\newblock In \emph{ICLR}, 2022.

\bibitem[MLT()]{MLT}
MLT.
\newblock Icdar 2019 robust reading challenge on multi-lingual scene text detection and recognition.
\newblock https://rrc.cvc.uab.es/?ch=15, 2019.

\bibitem[ModelScope(2023)]{DuGuangOCR}
ModelScope.
\newblock Duguangocr.
\newblock \url{https://modelscope.cn/models/damo/cv_convnextTiny_ocr-recognition-general_damo/summary}, 2023.

\bibitem[Mou et~al.(2023)Mou, Wang, Xie, Zhang, Qi, Shan, and Qie]{Mou_T2I_Corr23}
Chong Mou, Xintao Wang, Liangbin Xie, Jian Zhang, Zhongang Qi, Ying Shan, and Xiaohu Qie.
\newblock T2i-adapter: Learning adapters to dig out more controllable ability for text-to-image diffusion models.
\newblock \emph{arXiv preprint}, abs/2302.08453, 2023.

\bibitem[MTWI()]{MTWI}
MTWI.
\newblock Icpr 2018 challenge on multi-type web images.
\newblock https://tianchi.aliyun.com/dataset/137084, 2018.

\bibitem[Nichol \& Dhariwal(2021)Nichol and Dhariwal]{Nichol_ImprovedDDPM_ICML21}
Alexander~Quinn Nichol and Prafulla Dhariwal.
\newblock Improved denoising diffusion probabilistic models.
\newblock In \emph{ICML}, volume 139, pp.\  8162--8171, 2021.

\bibitem[PaddlePaddle(2023)]{PP-OCRv4}
PaddlePaddle.
\newblock Pp-ocrv4.
\newblock \url{https://github.com/PaddlePaddle/PaddleOCR/blob/release/2.7/doc/doc_ch/PP-OCRv4_introduction.md}, 2023.

\bibitem[Ramesh et~al.(2021)Ramesh, Pavlov, Goh, Gray, Voss, Radford, Chen, and Sutskever]{Ramesh_DALLE_icml21}
Aditya Ramesh, Mikhail Pavlov, Gabriel Goh, Scott Gray, Chelsea Voss, Alec Radford, Mark Chen, and Ilya Sutskever.
\newblock Zero-shot text-to-image generation.
\newblock In \emph{ICML}, volume 139, pp.\  8821--8831. {PMLR}, 2021.

\bibitem[Ramesh et~al.(2022)Ramesh, Dhariwal, Nichol, Chu, and Chen]{Ramesh_DALLE2_corr22}
Aditya Ramesh, Prafulla Dhariwal, Alex Nichol, Casey Chu, and Mark Chen.
\newblock Hierarchical text-conditional image generation with {CLIP} latents.
\newblock \emph{arXiv preprint}, abs/2204.06125, 2022.

\bibitem[RCTW()]{RCTW}
RCTW.
\newblock Icdar2017 competition on reading chinese text in the wild.
\newblock https://rctw.vlrlab.net/dataset, 2017.

\bibitem[ReCTS()]{ReCTS}
ReCTS.
\newblock Icdar 2019 robust reading challenge on reading chinese text on signboard.
\newblock https://rrc.cvc.uab.es/?ch=12, 2019.

\bibitem[Rodriguez et~al.(2023)Rodriguez, V{\'{a}}zquez, Laradji, Pedersoli, and Rodr{\'{\i}}guez]{Rodriguez_OCR-VQGAN_WACV23}
Juan~A. Rodriguez, David V{\'{a}}zquez, Issam~H. Laradji, Marco Pedersoli, and Pau Rodr{\'{\i}}guez.
\newblock {OCR-VQGAN:} taming text-within-image generation.
\newblock In \emph{WACV}, pp.\  3678--3687, 2023.

\bibitem[Rombach et~al.(2022)Rombach, Blattmann, Lorenz, Esser, and Ommer]{Rombach_LDM_CVPR22}
Robin Rombach, Andreas Blattmann, Dominik Lorenz, Patrick Esser, and Bj\"orn Ommer.
\newblock High-resolution image synthesis with latent diffusion models.
\newblock In \emph{CVPR}, pp.\  10684--10695, June 2022.

\bibitem[Saharia et~al.(2022)Saharia, Chan, Saxena, Li, Whang, Denton, Ghasemipour, Lopes, Ayan, Salimans, Ho, Fleet, and Norouzi]{Saharia_Imagen_NIPS22}
Chitwan Saharia, William Chan, Saurabh Saxena, Lala Li, Jay Whang, Emily~L. Denton, Seyed Kamyar~Seyed Ghasemipour, Raphael~Gontijo Lopes, Burcu~Karagol Ayan, Tim Salimans, Jonathan Ho, David~J. Fleet, and Mohammad Norouzi.
\newblock Photorealistic text-to-image diffusion models with deep language understanding.
\newblock In \emph{NeurIPS}, 2022.

\bibitem[Schuhmann et~al.(2021)Schuhmann, Vencu, Beaumont, Kaczmarczyk, Mullis, Katta, Coombes, Jitsev, and Komatsuzaki]{laion_400m_corr21}
Christoph Schuhmann, Richard Vencu, Romain Beaumont, Robert Kaczmarczyk, Clayton Mullis, Aarush Katta, Theo Coombes, Jenia Jitsev, and Aran Komatsuzaki.
\newblock {LAION-400M:} open dataset of clip-filtered 400 million image-text pairs.
\newblock \emph{CoRR}, abs/2111.02114, 2021.

\bibitem[Schuhmann et~al.(2022)Schuhmann, Beaumont, Vencu, Gordon, Wightman, Cherti, Coombes, Katta, Mullis, Wortsman, Schramowski, Kundurthy, Crowson, Schmidt, Kaczmarczyk, and Jitsev]{LAION-5b_NEURIPS2022}
Christoph Schuhmann, Romain Beaumont, Richard Vencu, Cade Gordon, Ross Wightman, Mehdi Cherti, Theo Coombes, Aarush Katta, Clayton Mullis, Mitchell Wortsman, Patrick Schramowski, Srivatsa Kundurthy, Katherine Crowson, Ludwig Schmidt, Robert Kaczmarczyk, and Jenia Jitsev.
\newblock Laion-5b: An open large-scale dataset for training next generation image-text models.
\newblock In S.~Koyejo, S.~Mohamed, A.~Agarwal, D.~Belgrave, K.~Cho, and A.~Oh (eds.), \emph{Advances in Neural Information Processing Systems}, volume~35, pp.\  25278--25294. Curran Associates, Inc., 2022.

\bibitem[Shi et~al.(2023)Shi, Xiong, Lin, and Jung]{Shi_InstantBooth_Corr23}
Jing Shi, Wei Xiong, Zhe Lin, and Hyun~Joon Jung.
\newblock Instantbooth: Personalized text-to-image generation without test-time finetuning.
\newblock \emph{arXiv preprint}, abs/2304.03411, 2023.

\bibitem[Song et~al.(2021)Song, Sohl{-}Dickstein, Kingma, Kumar, Ermon, and Poole]{Song_SDE_ICLR21}
Yang Song, Jascha Sohl{-}Dickstein, Diederik~P. Kingma, Abhishek Kumar, Stefano Ermon, and Ben Poole.
\newblock Score-based generative modeling through stochastic differential equations.
\newblock In \emph{ICLR}, 2021.

\bibitem[Wei et~al.(2023)Wei, Zhang, Ji, Bai, Zhang, and Zuo]{Wei_ELITE_Corr23}
Yuxiang Wei, Yabo Zhang, Zhilong Ji, Jinfeng Bai, Lei Zhang, and Wangmeng Zuo.
\newblock {ELITE:} encoding visual concepts into textual embeddings for customized text-to-image generation.
\newblock \emph{arXiv preprint}, abs/2302.13848, 2023.

\bibitem[Yang et~al.(2023)Yang, Gui, Yuan, Ding, Hu, and Chen]{Yang_GlyphControl_Corr23}
Yukang Yang, Dongnan Gui, Yuhui Yuan, Haisong Ding, Han Hu, and Kai Chen.
\newblock Glyphcontrol: Glyph conditional control for visual text generation.
\newblock \emph{arXiv preprint}, abs/2305.18259, 2023.

\bibitem[Zhang \& Agrawala(2023)Zhang and Agrawala]{Zhang_ControlNet_Corr23}
Lvmin Zhang and Maneesh Agrawala.
\newblock Adding conditional control to text-to-image diffusion models.
\newblock \emph{arXiv preprint}, abs/2302.05543, 2023.

\end{thebibliography}
\bibliographystyle{iclr2024_conference}

\appendix
\section{Appendix}

\subsection{More Examples on the Flexibility of AnyText}
\label{app: flexibility}
\begin{figure}[htbp]
 \vspace{-1mm}
 \centering
 \includegraphics[width=0.8\textwidth]{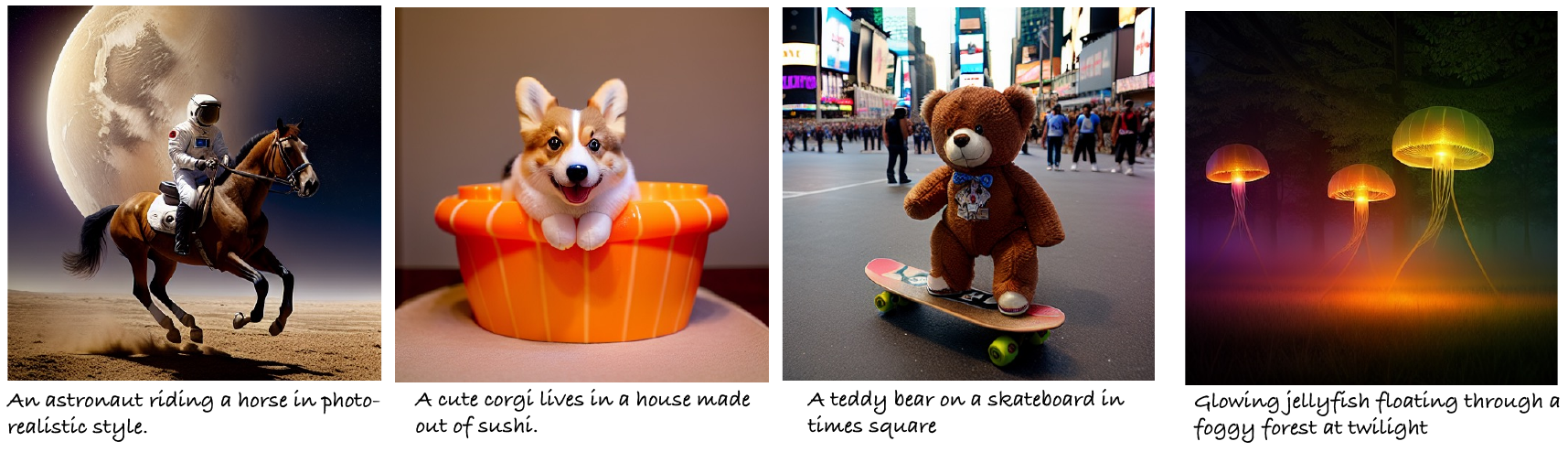}
 \vspace{-3mm}
 \caption{Examples of AnyText generating images without text.}
 \vspace{-0.15in}
 \label{fig:notext}
\end{figure}

\begin{figure}[htbp]
 \vspace{-1mm}
 \centering
 \includegraphics[width=0.8\textwidth]{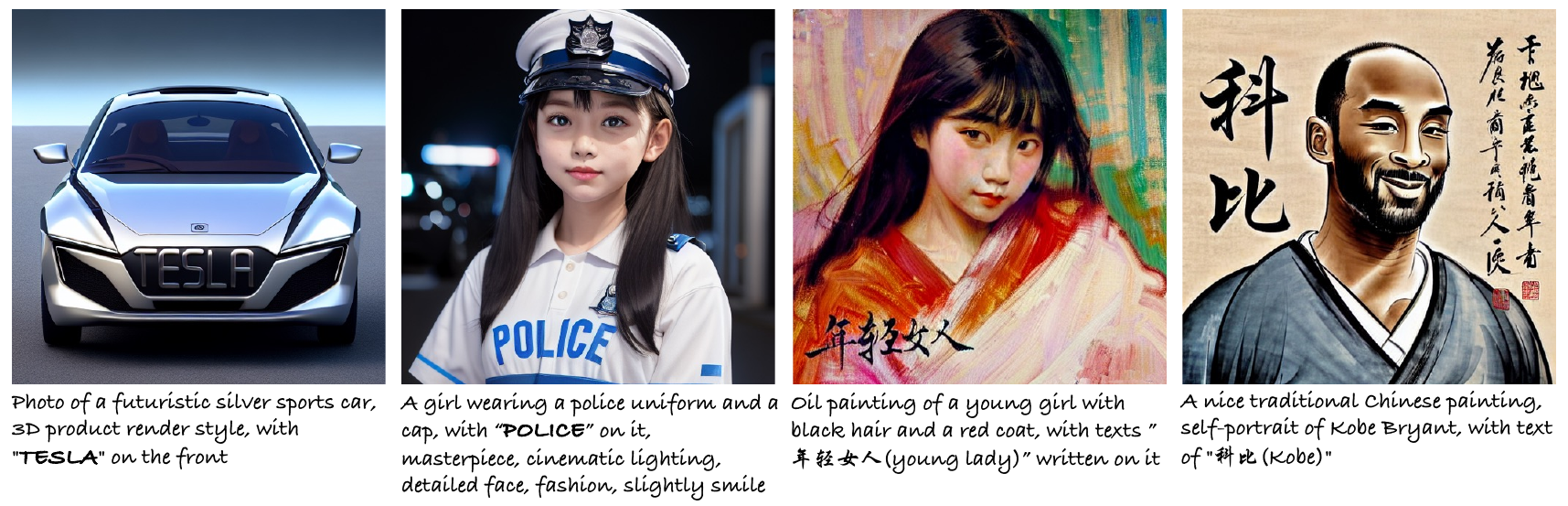}
 \vspace{-3mm}
 \caption{Examples of community models integrated with AnyText that can generate text.}
 \vspace{-0.15in}
 \label{fig:compatibility}
\end{figure}

By adopting the architecture of ControlNet, AnyText exhibits great flexibility. On one hand, the base model's conventional text-to-image generation capability is preserved as shown in Fig.~\ref{fig:notext}. On the other hand, the open-source community, such as HuggingFace and CivitAI, has developed a wide range of base models with diverse styles. By modularly binding AnyText, as depicted in Fig.~\ref{fig:compatibility}, these models(Oil Painting\footnote{https://civitai.com/models/20184}, Guoha Diffusion\footnote{https://civitai.com/models/33/guoha-diffusion}, Product Design\footnote{https://civitai.com/models/23893}, and Moon Film\footnote{https://civitai.com/models/43977}) can acquire the ability to generate text. All of these aspects highlight the highly versatile use cases of AnyText.

\subsection{Parameter Size and Computational Overhead of AnyText}
\label{app: param_size}
Our framework is implemented based on ControlNet. Despite the addition of some modules, it did not significantly increase the parameter size, as refered to Table~\ref{table:param_size}. We compared the computational overhead of both models using a batch size of 4 on a single Tesla V100, the inference time for ControlNet is 3476 ms/image, and for AnyText is 3512 ms/image.
\begin{table}
    \vspace{-1mm}
    \small
    \centering
    \caption{\small The Comparison of the parameter sizes of modules between ControlNet and AnyText.}
    \vspace{0.0in}
    \setlength{\tabcolsep}{1.35mm}{
    \begin{tabular}{l|c|c}
    \hline
    Modules   & ControlNet  & AnyText  \\ \hline
    UNet & 859M & 859M  \\
    VAE & 83.7M & 83.7M \\
    CLIP Text Encoder & 123M & 123M \\
    ControlNet & 361M & - \\
    TextControlNet & - & 360M \\
    Glyph Block & - & 0.35M \\
    Position Block & - & 0.04M \\
    Fuse Layer & - & 0.93M \\
    OCR Model & - & 2.6M  \\
    Linear Layer & - & 2.0M \\
    Total & 1426.7M & 1431.6M \\
    \end{tabular}}
    \label{table:param_size}
\end{table}

\subsection{More Details About Dataset Preparation}
\label{app: dataset}
\textbf{Filtering Rules}: We have established strict filtering rules to ensure the quality of training data. Taking the Chinese dataset Wukong~\cite{wukong_corr22} as an example, each original image undergoes the following filtering rules:
\begin{itemize}
    \item [\textbullet] Width or height of the image should be no less than 256.
    \item [\textbullet] Aspect ratio of the image should be between 0.67 and 1.5.
    \item [\textbullet] Area of the text region should not be less than 10\% of the entire image.
    \item [\textbullet] Number of text lines in the image should not exceed 8.
\end{itemize}
Next, for the remaining images, each text line is filtered based on the following rules:
\begin{itemize}
    \item [\textbullet] Height of the text should not be less than 30 pixels.
    \item [\textbullet] Score of OCR recognition of the text should be no lower than 0.7.
    \item [\textbullet] Content of the text should not be empty or consist solely of whitespace.
\end{itemize}
We implemented similar filtering rules on the LAION-400M~\cite{laion_400m_corr21} dataset, although the rules were more stringent due to the abundance of English data compared to Chinese. By implementing these rigorous filtering rules, we aim to ensure that the training data consists of high-quality images and accurately recognized text lines.

\textbf{Image Captions}: We regenerated caption for each image using BLIP-2~\cite{Li_BLIP2_Corr23} and removed specific placeholders $S_*$ (such as `*'). Then, we randomly selected one of the following statements and concatenated it to the caption:
\begin{itemize}
    \item [\textbullet] ``, content and position of the texts are "
    \item [\textbullet] ``, textual material depicted in the image are "
    \item [\textbullet] ``, texts that say "
    \item [\textbullet] ``, captions shown in the snapshot are "
    \item [\textbullet] ``, with the words of "
    \item [\textbullet] ``, that reads "
    \item [\textbullet] ``, the written materials on the picture: "
    \item [\textbullet] ``, these texts are written on it: "
    \item [\textbullet] ``, captions are "
    \item [\textbullet] ``, content of the text in the graphic is " 
\end{itemize}
Next, based on the number of text lines, the placeholders will be concatenated at the end to form the final textual description, such as: ``\textit{a button with an orange and white design on it, these texts are written on it: *, *, *, *}". After processing through the text embedding module, the embeddings corresponding to the placeholders will be replaced with embeddings of the text's glyph information.

\subsection{Stastic and Examples of AnyWord-3M}
\label{app: stastic}
\begin{figure}[htbp]
 \centering
 \includegraphics[width=1.0\textwidth]{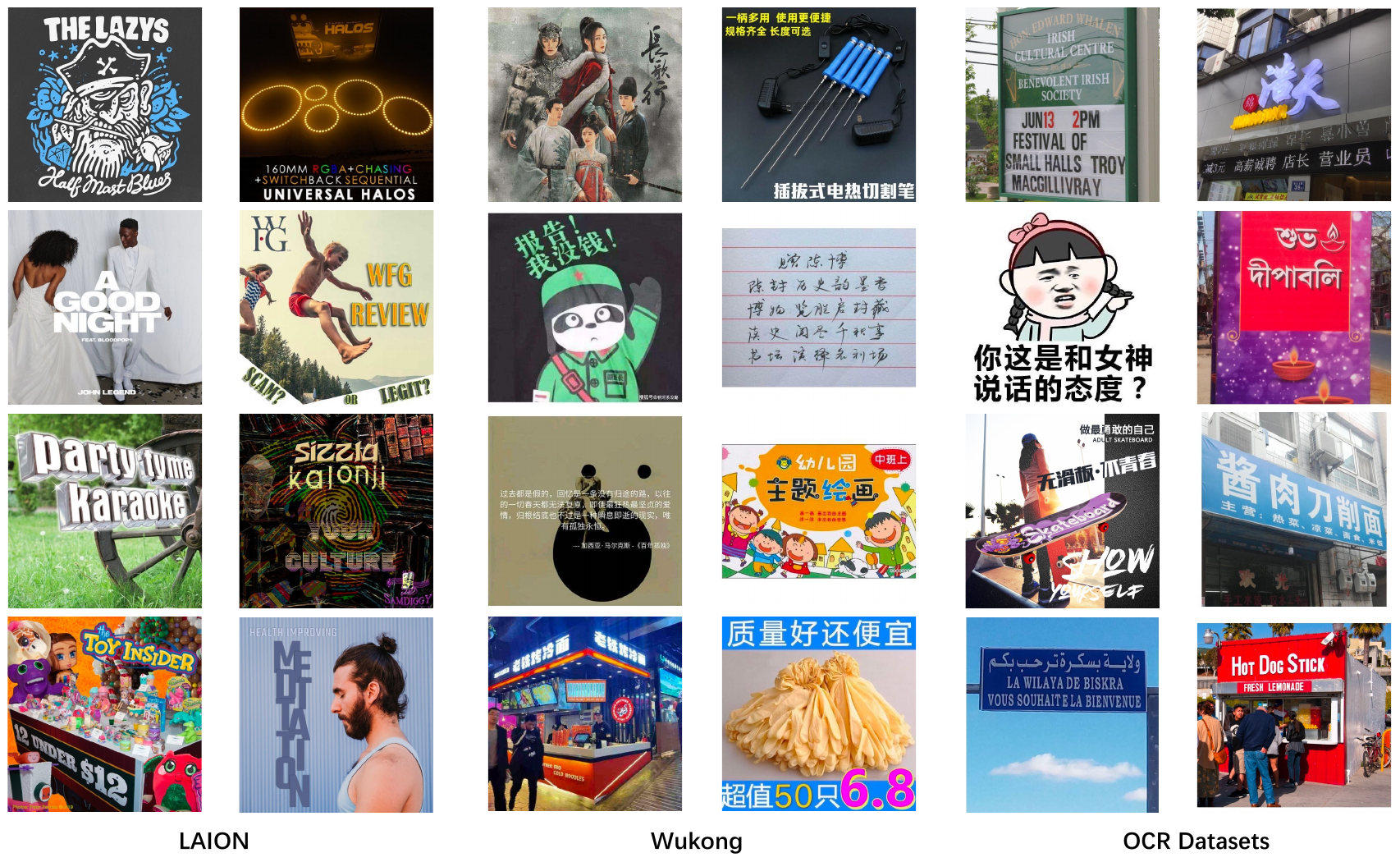}
 \caption{Examples of images from the AnyWord-3M dataset.}
 \label{fig:anyword-3m}
\end{figure}

In Table~\ref{table:dataset_stat1} and Table~\ref{table:dataset_stat2}, we provide detailed statistics on the composition of AnyWord-3M dataset. Additionally, in Fig.~\ref{fig:anyword-3m}, we present some example images from the dataset.

\begin{table}
    \vspace{-8mm}
    \small
    \centering
    \caption{\small Statistics of dataset size and line count in subsets of AnyWord-3M.}
    \vspace{0.0in}
    \setlength{\tabcolsep}{1.35mm}{
    \begin{tabular}{l|c|c|c|c|c}
    \hline
    Subsets   & image count  & image w/o text & line count & mean lines/img  & \#img $<=$5 lines  \\ \hline
    Wukong  & 1.54M & 0 & 3.23M & 2.10 & 1.51M, 98.1\%  \\
    LAION   & 1.39M & 0 & 5.75M & 4.13 & 1.03M, 74.0\%  \\
    OCR datasets  & 0.1M & 21.7K & 203.5K & 2.03 & 93.7K, 93.6\%  \\
\hline
    Total  & 3.03M & 21.7K & 9.18M & 3.03 & 2.64M, 86.9\% \\ \hline
    \end{tabular}}
    \label{table:dataset_stat1}
\end{table}

\begin{table}
    \vspace{-8mm}
    \small
    \centering
    \caption{\small Statistics of characters or words for different languages in AnyWord-3M.}
    \vspace{0.0in}
    \setlength{\tabcolsep}{1.35mm}{
    \begin{tabular}{l|c|c|c|c|c}
    \hline
    Lanuages   & line count  & chars/words count & unique chars/words & mean chars/line  & \#line $<=$20 chars  \\ \hline
    Chinese  & 2.90M & 15.09M & 5.9K & 5.20 & 2.89M, 99.5\%  \\
    English   & 6.27M & 6.35M & 695.2K & 5.42 & 6.25M, 99.7\%  \\
    Others  & 11.7K & 59.5K & 2.1K & 5.06 & 11.7K, 100\%  \\
\hline
    Total  & 9.18M & 21.50M & 703.3K & 5.35 & 9.15M, 99.6\% \\ \hline
    \end{tabular}}
    \label{table:dataset_stat2}
\end{table}

\subsection{Examples from AnyText-benchmark}
\label{app:eval-examples}
\begin{figure}[htbp]
 \centering
 \includegraphics[width=0.95\textwidth]{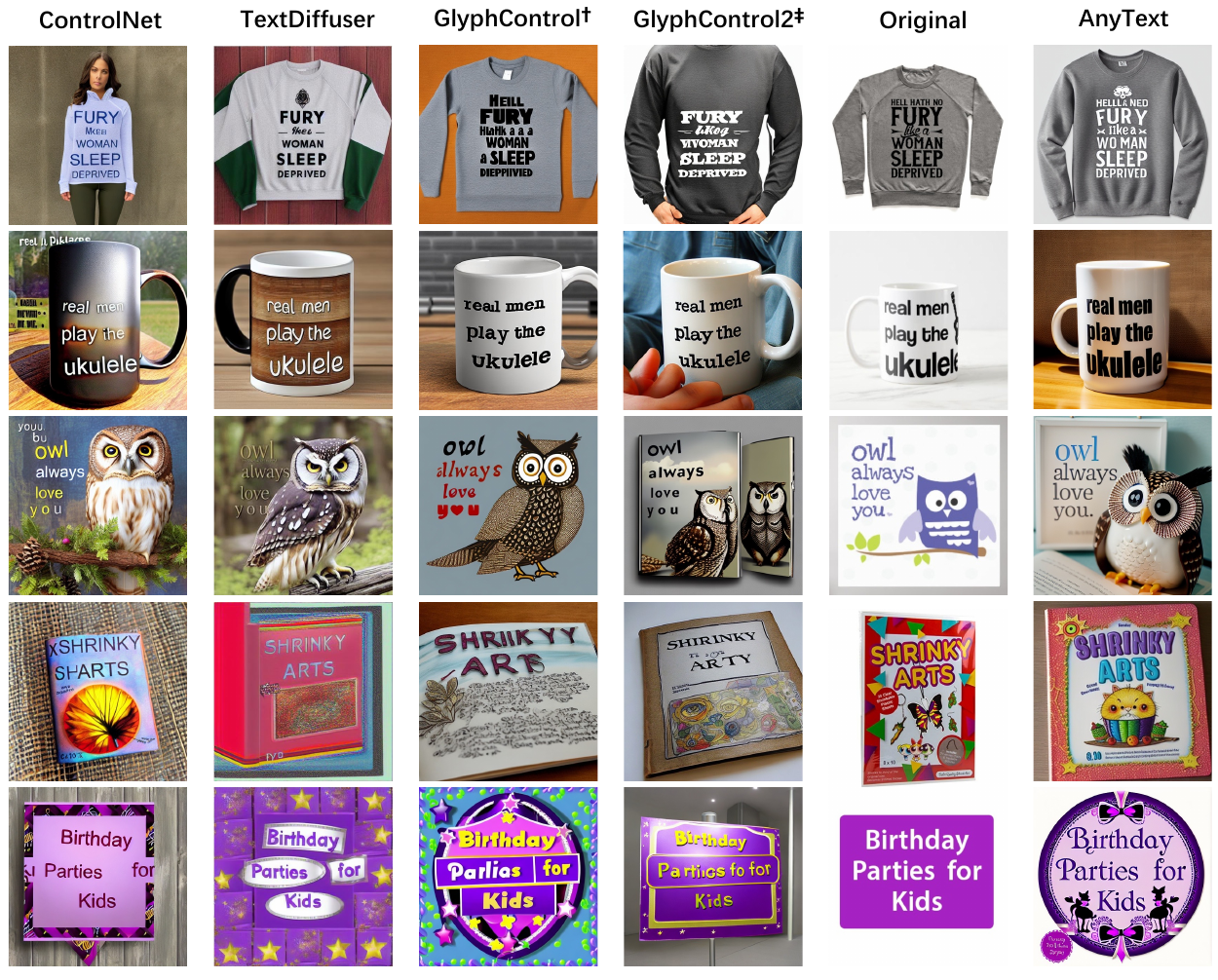}
 \vspace{-5mm}
 \caption{Examples of images in English from the AnyText-benchmark. \dag is trained on LAION-Glyph-10M, and \ddag is fine-tuned on TextCaps-5k.}
 \vspace{-5mm}
 \label{fig:eval-laion}
\end{figure}

\begin{figure}[htbp]
 \centering
 \includegraphics[width=0.95\textwidth]{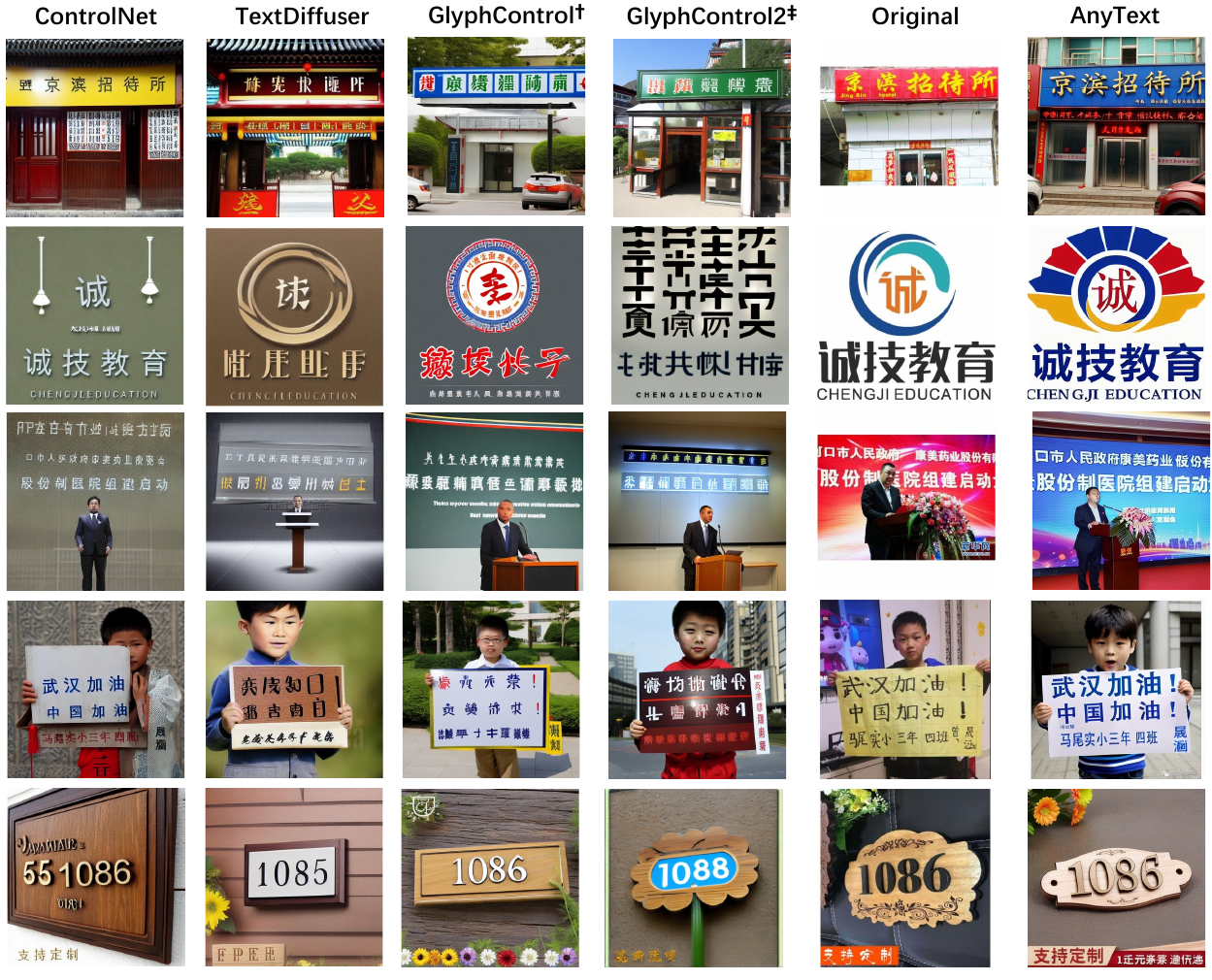}
 \vspace{-5mm}
 \caption{Examples of images in Chinese from the AnyText-benchmark. \dag is trained on LAION-Glyph-10M, and \ddag is fine-tuned on TextCaps-5k.}
 \vspace{-5mm}
 \label{fig:eval-wukong}
\end{figure}

We selected some images from the generated image of AnyText-benchmark evaluation set. The English and Chinese examples can be seen in Fig.~\ref{fig:eval-laion} and Fig.~\ref{fig:eval-wukong}, respectively. All the example images were generated using the same fixed random seed, as well as the same positive prompt (\textit{``best quality, extremely detailed"}) and negative prompt (\textit{``longbody, lowres, bad anatomy, bad hands, missing fingers, extra digit, fewer digits, cropped, worst quality, low quality, watermark"}).

\subsection{More Examples of AnyText}
\label{app: more_examples}
\begin{figure}[htbp]
 \centering
 \includegraphics[width=1.0\textwidth]{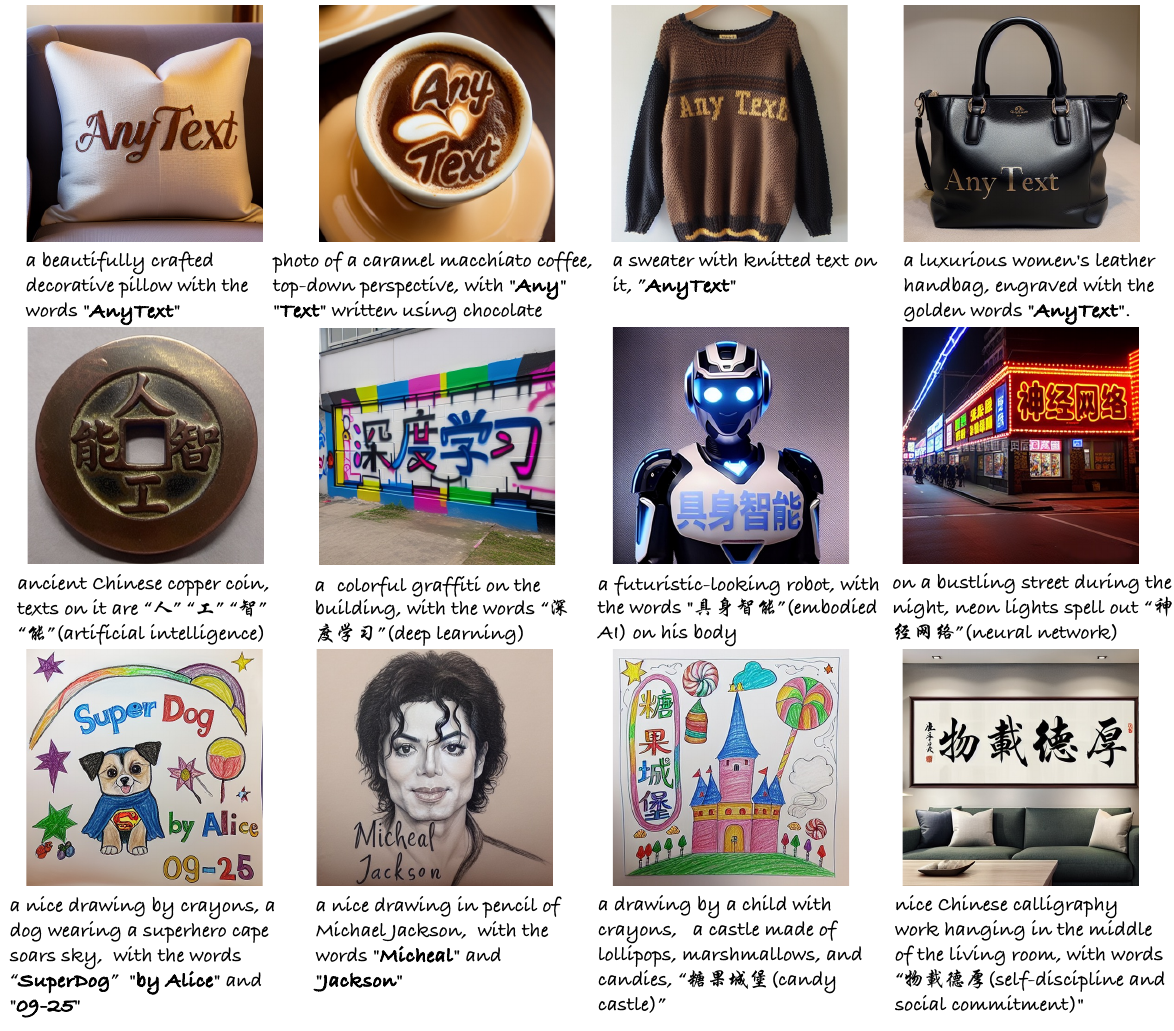}
 \caption{More examples of AnyText in text generation.}
 \label{fig:more_examples_generate}
\end{figure}

\begin{figure}[htbp]
 \centering
 \includegraphics[width=1.0\textwidth]{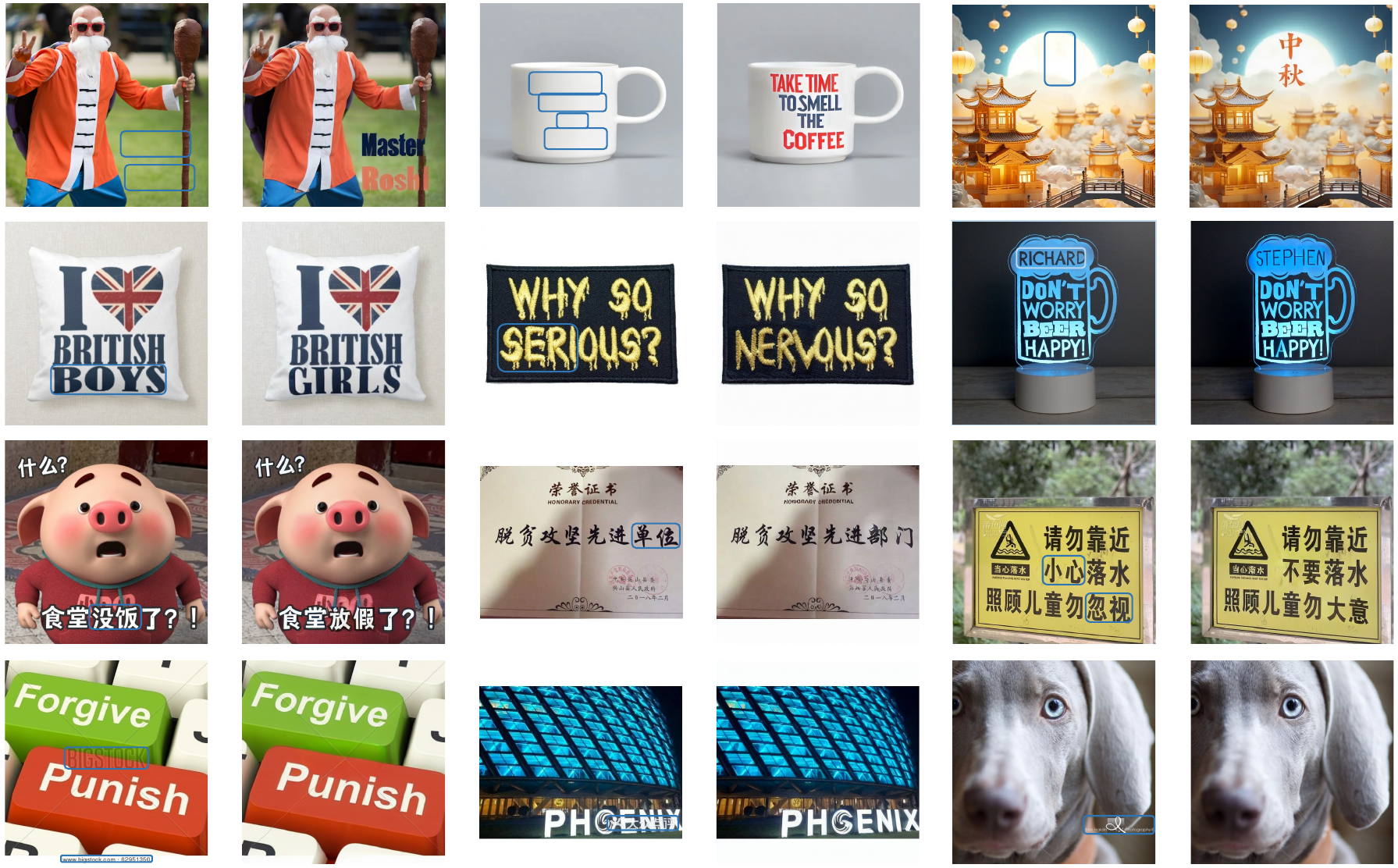}
 \caption{More examples of AnyText in text editing.}
 \label{fig:more_examples_edit}
\end{figure}

We present additional examples in the text generation (see Fig.~\ref{fig:more_examples_generate}) and text editing (see Fig.~\ref{fig:more_examples_edit}).

\subsection{Implementation Details of AnyText-v1.1}
\label{app: details_v1.1}
We randomly sampled some text lines in AnyWord-3M and observed that the OCR annotation error rate for the English portion was 2.4\%, while for the Chinese portion was significantly higher at 11.8\%. So, for the Chinese part, we used the latest PP-OCRv4~\cite{PP-OCRv4} to regenerate the OCR annotations; and for the English part, we found that the annotations from MARIO-LAION by TextDiffuser~\cite{Chen_TextDiffuser_Corr23} performed slightly better on some severely deformed English sentences compared to PP-OCRv4, so we replaced them with their annotations. Additionally, we slightly increased some data and making the ratio of the English and Chinese was approximately 1:1. This updated dataset was labeled as v1.1, as shown in Table~\ref{table:datasetv1.1}. In addition, we added a ``wm\_score" label for each image to indicate the probability of containing a watermark, which is used to filter images with watermarks in the final stage of model training.

\begin{table}
    \vspace{-4mm}
    \small
    \centering
    \caption{\small Changes in the data scale of AnyWord-3M.}
    \vspace{0.0in}
    \setlength{\tabcolsep}{1.35mm}{
    \begin{tabular}{c|c|c|c|c}
    \hline
    Version   & Wukong  & LAION & Total & wm\_score$<$0.5  \\ \hline
    v1.0  & 1.54M & 1.39M & 3.03M & --  \\
    v1.1   & 1.71M & 1.72M & 3.53M & 2.95M  \\  \hline
    \end{tabular}}
    \label{table:datasetv1.1}
\end{table}

Furthermore, we proposed a method called ``inv\_mask". For each text line in the image, if its recognition score is too low, or the text is too small, or if it is not among the 5 randomly chosen text lines, it will be marked as invalid. Subsequently, these invalid text lines are combined to form a mask, and during training, the loss in the corresponding area will be set to 0. This straightforward approach can bring significant improvement in OCR metrics, as illustrated in ~\ref{table:quanti_res}, and also notably reduced the ratio of pseudo-text areas in the background, as depicted in ~\ref{table:wm_pt}.

\begin{table}
    \vspace{-4mm}
    \small
    \centering
    \caption{\small Improvement of watermark and pseudo-text of AnyText, tested on private model and dataset. }
    \vspace{0.0in}
    \setlength{\tabcolsep}{1.5mm}{
    \begin{tabular}{c|c|c|c|c}
    \hline
    \multirow{2}{*}{Version}   & \multicolumn{2}{c|}{English}  & \multicolumn{2}{c}{Chinese}  \\
    \cline{2-5}  & watermark & pseudo-text & watermark & pseudo-text \\ \hline
    v1.0   & 6.9\% & 119.0\% & 24.7\% & 166.8\%  \\
    v1.1  & 0.4\% & 110.9\% & 2.9\% & 124.4\%  \\
    \hline
\end{tabular}}
    \vspace{-0.15in}
    \label{table:wm_pt}
\end{table}

The v1.0 model underwent further fine-tuning on the AnyWord-3M v1.1 dataset for 5 epochs, with the last 2 epochs using data of wm\_score $<$ 0.5, filtering out approximately 25\% Chinese data and 8\% English data. The parameter $\lambda$ for perceptual loss was set to 0.003, as it yielded favorable results for both Chinese and English on the v1.1 dataset. Subsequent to the training process, the original SD1.5 base model was replaced with a community model \textit{Realistic\_Vision}\footnote{https://huggingface.co/SG161222/Realistic\_Vision\_V5.0\_noVAE}, which makes the generated images more aesthetically appealing. The final model was labeled as version v1.1.

\end{document}